%% file: main.tex
\documentclass[10pt,journal,compsoc]{IEEEtran}
\usepackage{amsmath,amsfonts}
\usepackage{algorithmic}
\usepackage{algorithm}
\usepackage{array}
\usepackage[caption=false,font=normalsize,labelfont=sf,textfont=sf]{subfig}
\usepackage{textcomp}
\usepackage{ulem}
\usepackage{stfloats}
\usepackage{url}
\usepackage{verbatim}
\usepackage{graphicx}
\usepackage{cite}
\usepackage{graphicx}
\usepackage{amsmath}
\usepackage{amssymb}
\usepackage{booktabs}
\usepackage{multicol}
\usepackage{multirow}
\usepackage{diagbox}
\usepackage{graphicx}
\usepackage{ulem}

\usepackage{xcolor}         
\usepackage{adjustbox}
\usepackage{ulem}

\usepackage{url}
\usepackage{wrapfig}
\usepackage{listings}
\usepackage{algorithm}
\usepackage[capitalize]{cleveref}
\crefname{section}{Sec.}{Secs.}
\Crefname{section}{Section}{Sections}
\Crefname{table}{Table}{Tables}
\crefname{table}{Tab.}{Tabs.}
\newcommand{\tablestyle}[2]{\setlength{\tabcolsep}{#1}\renewcommand{\arraystretch}{#2}\centering\footnotesize}
\newlength\savewidth\newcommand\shline{\noalign{\global\savewidth\arrayrulewidth
  \global\arrayrulewidth 1pt}\hline\noalign{\global\arrayrulewidth\savewidth}}

\usepackage{float} 
\usepackage{color}
\usepackage{colortbl}

\newcommand{\wenlj}[1]{{\color{black}{#1}}}
\newcommand{\wenljx}[1]{{\color{black}{#1}}}

\definecolor{CUHKmiddle}{RGB}{144,44,144}
\usepackage{tcolorbox}
\tcbuselibrary{skins,breakable}
    {\endtcolorbox}

\begin{document}

\title{MVEB: Self-Supervised Learning with Multi-View Entropy Bottleneck}

\author{Liangjian  Wen, 
        Xiasi Wang,
        Jianzhuang Liu,
        Zenglin Xu 
\IEEEcompsocitemizethanks{
\IEEEcompsocthanksitem Corresponding to Zenglin Xu
\IEEEcompsocthanksitem L. Wen is with the School of Computing and Artificial Intelligence,and Research Institute for Digital Economy and Interdisciplinary Sciences,
Southwestern University of Finance and Economics, Chengdu, China. \\E-mail: wlj6816@gmail.com

\IEEEcompsocthanksitem  X. Wang is with the Hong Kong University of Science and Technology, Hong Kong, China. E-mail: xwangfy@connect.ust.hk
\IEEEcompsocthanksitem J. Liu is with  Shenzhen Institute of Advanced Technology, Shenzhen, China. E-mail: jz.liu@siat.ac.cn

\IEEEcompsocthanksitem  Z. Xu is with  the Harbin Institute of Technology Shenzhen, Shenzhen, China, and the Pengcheng Laboratory, Shenzhen, China. E-mail: zenglin@gmail.com
\IEEEcompsocthanksitem This work was partially supported by an Open Research Project of Zhejiang Lab (NO.2022RC0AB04), a Major Key Project of PCL (No. PCL2023A09), and a key program of fundamental research from Shenzhen Science and Technology Innovation Commission (No. JCYJ20200109113403826)
}       

}

\markboth{Journal of \LaTeX\ Class Files,~Vol.~14, No.~8, August~2021}%
{Shell \MakeLowercase{\textit{et al.}}: A Sample Article Using IEEEtran.cls for IEEE Journals}

\input{doc/abstract}

\maketitle

\input{doc/intro}
\input{doc/related_work}

\input{doc/prelim}

\input{doc/approach}

\input{doc/exp}

\input{doc/conclusion}


\bibliographystyle{IEEEtran}
\bibliography{egbib}
\vspace{-30pt}
\begin{IEEEbiography}[{\includegraphics[width=1in,height=1.25in,clip,keepaspectratio]{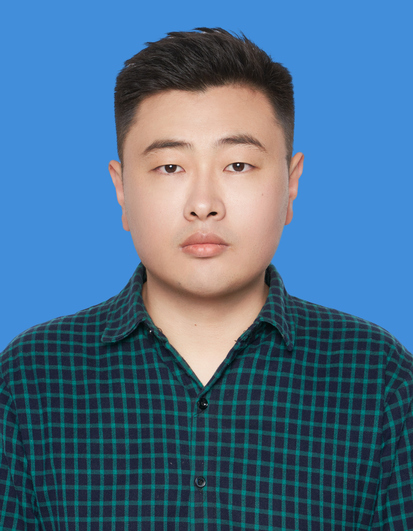}}]{Liangjian Wen} is  currently an assistant professor with the School of Computing and Artificial Intelligence,
Southwestern University of Finance and Economics, Chengdu, China.  I obtained my Ph.D. degree from School of Computer Science and Engineering, University of Electronic Science and Technology of China (UESTC) in July. 2021.
His research interests include machine learning, deep learning, representation learning, and self-supervised learning.
\end{IEEEbiography}
\vspace{-35pt}
\begin{IEEEbiography}[{\includegraphics[width=1in,height=1.25in,clip,keepaspectratio]{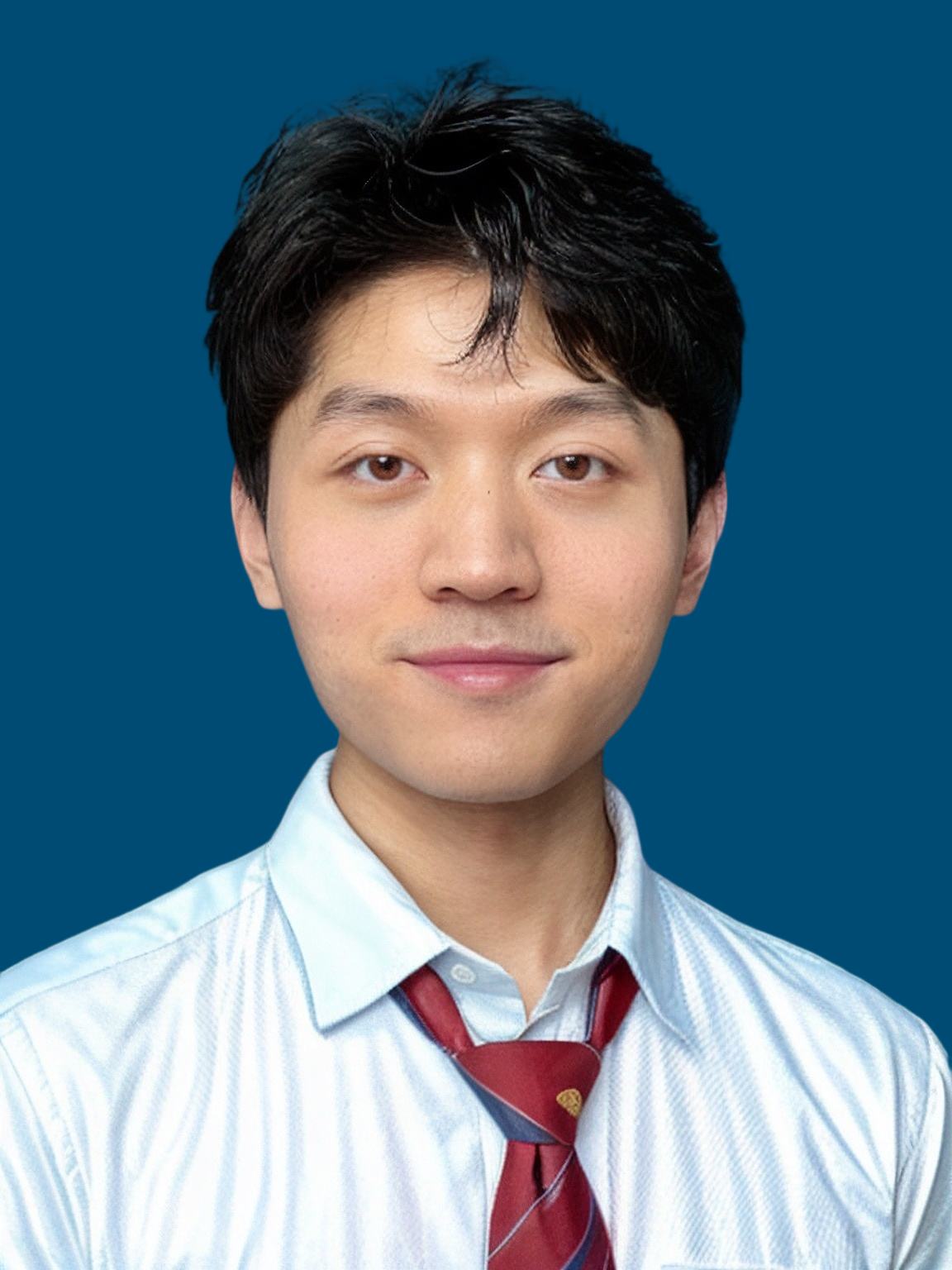}}]{Xiasi Wang} is a Ph.D. student majoring in IIP (Data Science and Analytics) at Academy of Interdisciplinary Studies, the Hong Kong University of Science and Technology, and he is supervised by Prof. Yuan Yao and Prof. Nevin L. Zhang. He received the Bachelor degree of Statistics from the School of Management at University of Science and Technology of China in 2020. His research interest lies in representation learning.
\end{IEEEbiography}
\vspace{-35pt}
\begin{IEEEbiography}[{\includegraphics[width=1in,height=1.25in,clip,keepaspectratio]{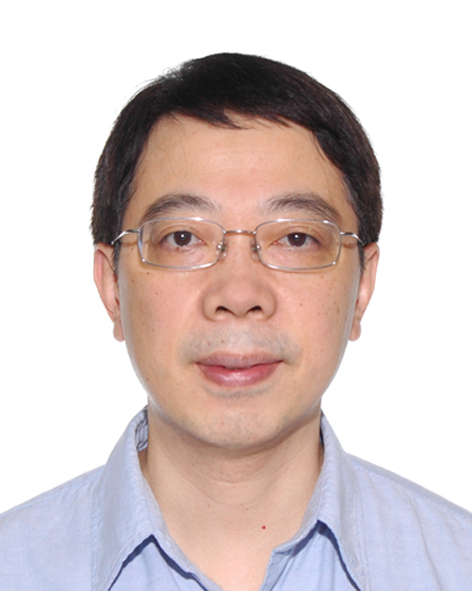}}]{Jianzhuang Liu}  (Senior Member, IEEE) received the PhD degree in computer vision from The Chinese University of Hong Kong, in 1997. From 1998 to 2000, he was a research fellow with Nanyang Technological University, Singapore. From 2000 to 2012, he was a post-doctoral fellow, an assistant professor, and an adjunct associate professor with The Chinese University of Hong Kong, Hong Kong. In 2011, he joined the Shenzhen Institute of Advanced Technology, University of Chinese Academy of Sciences, Shenzhen, China, as a professor. He was a principal researcher in Huawei Company from 2012 to 2023. He has authored more than 200 papers in the areas of computer vision, image processing, deep learning, and AIGC.
\end{IEEEbiography}
\vspace{-35pt}
\begin{IEEEbiography}[{\includegraphics[width=1in,height=1.25in,clip,keepaspectratio]{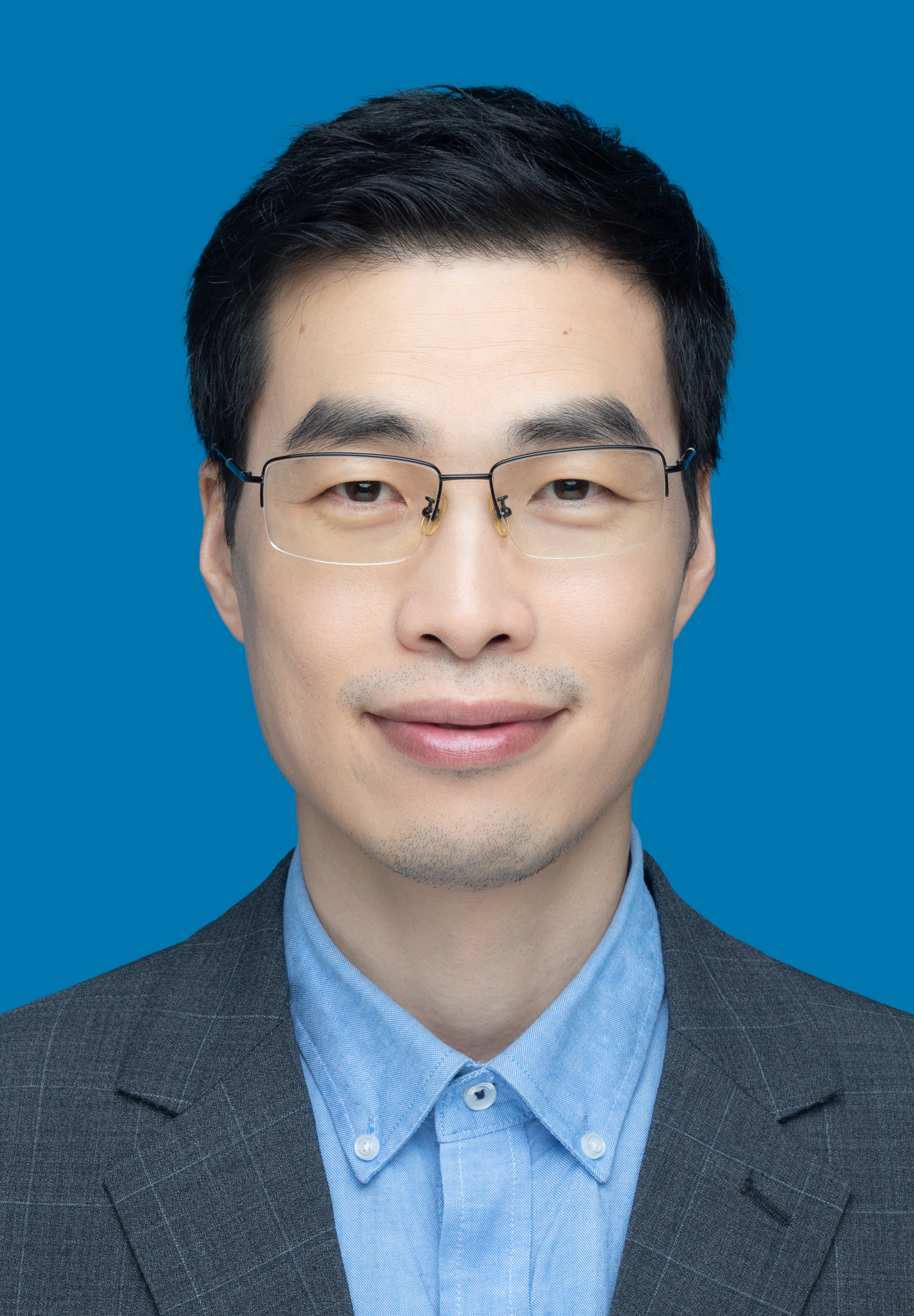}}]{Zenglin Xu} received a Ph.D. degree in computer science and engineering from the Chinese University of Hong Kong. He is currently a full professor at the Harbin Institute of Technology Shenzhen, and also affiliated with Peng Cheng Lab. He has been with Michigan State University, the Cluster of Excellence at Saarland University and the Max Planck Institute for Informatics, Purdue University, and the University of Electronic Science and Technology of China. Dr. Xu's research interests include machine learning and its applications in computer vision, health informatics, and natural language processing. He is the recipient of the outstanding student paper honorable mention at AAAI 2015, the best student paper runner up at ACML 2016, and the 2016 young researcher award from APNNS. He serves as an associate editor of Neural Networks. He is currently the vice president for education in the International Neural Network Society.
\end{IEEEbiography}
\clearpage

\newpage
\input{doc/sup.tex}

\end{document}

%% file: doc/abstract.tex
\IEEEtitleabstractindextext{
\begin{abstract}
Self-supervised learning aims to learn representation that can be effectively generalized to downstream tasks. Many self-supervised approaches regard two views of an image as both the input and the self-supervised signals, assuming that either view contains the same task-relevant information and the shared information is (approximately) sufficient for predicting downstream tasks. Recent studies  show that discarding superfluous information not  shared between the views can improve generalization. Hence, the ideal representation is sufficient for downstream tasks and contains minimal superfluous information, termed  minimal sufficient representation. One can learn this representation  by maximizing the mutual information between the representation and the supervised view while eliminating superfluous information. Nevertheless, the computation of mutual information is notoriously intractable. In this work, we propose an objective  termed  multi-view entropy bottleneck (MVEB) to learn minimal sufficient representation effectively. MVEB simplifies the minimal sufficient learning to maximizing both the agreement between the embeddings of two views and the differential entropy of the embedding distribution.  Our experiments confirm that MVEB significantly improves performance. For example, it achieves top-1 accuracy of 76.9\% on ImageNet with a vanilla ResNet-50 backbone on linear evaluation. To the best of our knowledge, this is the new state-of-the-art result with ResNet-50.
\end{abstract}
}

%% file: doc/intro.tex
\section{Introduction}
\IEEEPARstart{S}elf-supervised learning (SSL) has achieved significant progress in learning representation to generalize well to broad  downstream tasks. 
Many state-of-the-art SSL approaches  maximize the agreement between the embeddings of two views of an image from a multi-view perspective. 
These works are based on Siamese networks and employ different methods to deal with the collapse problem during representation learning. 
For example, contrastive learning~\cite{DBLP:conf/cvpr/WuXYL18,DBLP:conf/icml/ChenK0H20,DBLP:conf/cvpr/He0WXG20,9873966,9792378} utilizes  negative samples to separate features of different images to avoid collapse.  Asymmetric network methods~\cite{DBLP:conf/nips/GrillSATRBDPGAP20,chen2021exploring,caron2021emerging} introduce  a predictor network and a
momentum encoder (or a stop-gradient operation) to prevent  collapse without negative samples. In addition, feature
decorrelation methods~\cite{zbontar2021barlow,DBLP:conf/iclr/BardesPL22} avoid collapse by reducing the redundancy among  the feature dimensions. Empirical results of these SSL works show competitive performance on multiple visual tasks compared with supervised learning methods.

\begin{figure}
\centering 
\includegraphics[width=0.48\textwidth]{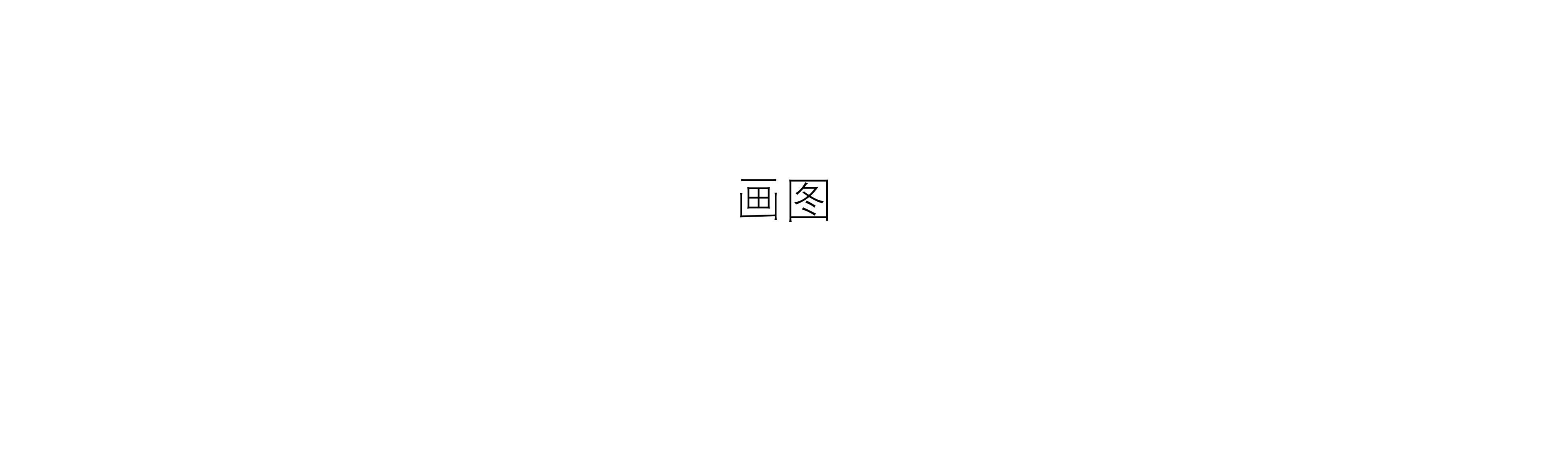}
\caption{Illustration of the sufficient and minimal representation in the unsupervised multi-view setting. The common assumption in multi-view learning is that the information $I(\mathbf{v_1};\mathbf{v_2})$ shared between view $\mathbf{v_1}$ and view $\mathbf{v_2}$ is sufficient for the prediction of downstream tasks~\cite{DBLP:conf/colt/SridharanK08}. When $I\left(\mathbf{z_1}; \mathbf{v_2}\right)=I\left(\mathbf{v_1}; \mathbf{v_2}\right)$, $\mathbf{z_1}$ (denoted by the dotted line in the above figure)  contains all the task-relevant information shared between the two views (left). Hence, $\mathbf{z_1}$ is a sufficient representation.   If all superfluous information is eliminated, i.e., $I\left(\mathbf{z_1}; \mathbf{v_1}\right)=I\left(\mathbf{v_1}; \mathbf{v_2}\right)$ (right), $\mathbf{z_1}$ is the minimal sufficient representation.}
\vspace{-1.5em}
\label{figure:minimalsufficient}
\end{figure}

From the multi-view perspective, self-supervised approaches often consider two image views as the input and the self-supervised signals for each other. One can assume that  either view is (approximately) sufficient for the prediction of downstream tasks and  contains the same task-relevant information in multi-view learning~\cite{DBLP:conf/colt/SridharanK08}. 
This suggests that  different views of an image should not affect the prediction for downstream tasks. 
Similar to the role of  labels in supervised learning, two views as the mutual self-supervised signals are adapted to extract task-relevant information based on Siamese networks. If the learned representation contains the same task-relevant information shared between the two views,
 it is sufficient for downstream tasks. Moreover, superfluous information is identified as  that not shared by both views.
As shown in \cite{DBLP:conf/iclr/Federici0FKA20} and \cite{DBLP:conf/iclr/Tsai0SM21}, discarding superfluous information  can improve
the generalization of the learned representation for downstream tasks.
Hence, the ideal representation is sufficient for downstream tasks and contains minimal superfluous information. 
In light of the information bottleneck principle~\cite{tishby2000information}, 
the minimal sufficient representation is defined  in the unsupervised setting, as shown in Fig.~\ref{figure:minimalsufficient}. One can learn  the minimal sufficient representation by minimizing the mutual information between the extracted feature and its input view conditioned on  the other supervised view while  maximizing the mutual information between the extracted  feature and the  supervised view. This is termed the multi-view information bottleneck~\cite{DBLP:conf/iclr/Federici0FKA20}. 
However, the computation of mutual information is notoriously intractable. Although the variational method \cite{DBLP:conf/iclr/Federici0FKA20} can be introduced to overcome the intractability, \cite{wang2022rethinking} shows this cannot improve the performance for downstream tasks much compared with SimCLR~\cite{DBLP:conf/icml/ChenK0H20}.
Learning the minimal sufficient representation effectively in self-supervised representation learning is still challenging. 

To address this  problem, we propose a new objective function, the multi-view entropy bottleneck (MVEB), to learn the minimal sufficient representation.  
Our method can learn task-relevant information and eliminate
superfluous  information, which is  related to the multi-view information bottleneck. 
MVEB simplifies the minimal sufficient learning to  the process of  maximizing both the agreement between the embeddings of two  views of an image and the differential entropy of the  embedding distribution. Moreover, it can be directly applied to Siamese networks without  modification of the network structure and other complex designs. 

However, it is  intractable to compute the differential entropy of the embedding distribution since it  is unknown. We propose a score-based entropy estimator with the von Mises-Fisher kernel~\cite{1999Directional} to approximate the gradient of the differential entropy with model parameters, such that we can directly use the gradient approximation with model
parameters for backpropagation to maximize the differential entropy. It can increase the uniformity of the embeddings efficiently. Moreover, this formulation does not require a large batch size or a memory bank.

We empirically demonstrate that MVEB significantly improves the generalization of the learned representation for downstream tasks. 
Our main contributions are summarized as follows:
\begin{itemize}

\item We propose MVEB to learn the minimal sufficient representation in the unsupervised multi-view setting. It can be directly applied to Siamese networks without modification of the network structure
and other complex designs.

\item We present a score-based entropy estimator with the von Mises-Fisher kernel to approximate the gradient of the differential entropy of the embedding distribution w.r.t.
model parameters. This estimator can be used to maximize the differential entropy to increase uniformity.

\item  \wenljx{We first analyze contrastive learning (e.g., SimCLR and MOCO), asymmetric network methods (e.g., BYOL and SimSiam), and feature decorrelation methods (e.g., Barlow Twins and  VICReg) from learning the  minimal sufficient representation. Based on MVEB, we argue that these methods also try to learn the minimal sufficient representation by optimizing  alignment and uniformity.}

\item 
Comprehensive experiments are conducted to show the superior performance of MVEB. For example, it achieves top-1 accuracy of 76.9\% on ImageNet with a vanilla ResNet-50 backbone with a single-layer classifier fine-tuned. To the best of our knowledge,  this is the new  state-of-the-art result with ResNet-50.
\end{itemize}




%% file: doc/related_work.tex
\section{Related Work}

Self-supervised learning (SSL) learns representation by defining a pretext task without annotation. Pretext tasks are the core of SSL to provide supervision signals to mine the data structure. 
Recently, contrastive learning~\cite{ DBLP:conf/icml/ChenK0H20,DBLP:conf/cvpr/He0WXG20} makes promising progress and has reduced the gap with supervised learning on many computer vision benchmarks. It defines instance classification as the pretext task~\cite{DBLP:journals/pami/DosovitskiyFSRB16}. Specifically, each image is considered as one class and is discriminated  invariant to its own distortions and different from other images. \cite{DBLP:conf/cvpr/WuXYL18} proposes to use the InfoNCE loss to recognize  images by contrasting them with other images.  
\cite{ DBLP:conf/icml/ChenK0H20} and  \cite{DBLP:conf/cvpr/He0WXG20} employ Siamese networks to improve the performance. 
However, a large number of images as negative samples are needed to contrast the positive sample. 
It requires a large batch size or a memory bank. Clustering-based approaches~\cite{DBLP:conf/eccv/CaronBJD18,DBLP:conf/nips/CaronMMGBJ20} as the variants of contrastive learning keep consistency between cluster assignments for different views of images. 
Contrastive learning methods naturally align the embeddings of positive samples and separate the  embeddings of negative samples to achieve uniformity of representation~\cite{DBLP:conf/icml/0001I20}.
Our MVEB can be simplified to maximizing the alignment between view embeddings and the differential entropy of the embedding distribution.  Maximizing the differential entropy can also improve uniformity without the need for a large bath size or memory bank.

Recent works can produce high-quality representation without negative samples in contrastive learning.
Asymmetric network methods can learn representation by maximizing the agreement between the view embeddings based on Siamese networks. 
They rely on  asymmetric network architecture to prevent collapse. BYOL~\cite{DBLP:conf/nips/GrillSATRBDPGAP20}  introduces a predictor network for the online branch and a momentum encoder. SimSiam~\cite{chen2021exploring} adopts a stop-gradient operation for the target branch and a predictor network to avoid collapse. DINO~\cite{caron2021emerging} combines  the momentum encoder with  self-distillation on the Transformer backbone. However, these methods are not well understood, and hard to interpret their architectural tricks~\cite{DBLP:conf/iclr/BardesPL22}. 
Feature decorrelation methods~\cite{zbontar2021barlow,DBLP:conf/iclr/BardesPL22,DBLP:conf/iclr/ZhangZYZY22,DBLP:conf/icml/ErmolovSSS21} avoid collapse by reducing the redundancy among the feature dimensions.

From the multi-view perspective, \cite{DBLP:conf/iclr/Tsai0SM21} proposes an information-theoretical framework
to understand and explain the success of SSL.
Specifically,
two   views of an image can be considered as the input and the self-supervised signals for each other.  \cite{DBLP:conf/iclr/Federici0FKA20} proposes the  
multi-view information bottleneck for unsupervised multi-view learning to learn the minimal sufficient representation. However, the computation of mutual information is notoriously intractable. Our MVEB is  related to the multi-view information bottleneck and can be effectively applied to  Siamese networks to learn the minimal sufficient representation. \cite{wang2022rethinking} considers that the minimal sufficient representation   gives rise to the degradation of performance  for downstream tasks since not all task-relevant information is shared between views. Since either view is assumed (approximately) sufficient for downstream tasks for general data augmentation used in self-supervised learning~\cite{DBLP:conf/iclr/Tsai0SM21},
we argue that the non-shared task-relevant information between views can be ignored. Our  experiment results also  verify that the minimal sufficient representation can improve the generalization for downstream tasks.

%% file: doc/prelim.tex
\wenlj{\section{Preliminary: Minimal Sufficient Representation}}

\begin{figure*}
\centering 
\includegraphics[width=1\textwidth]{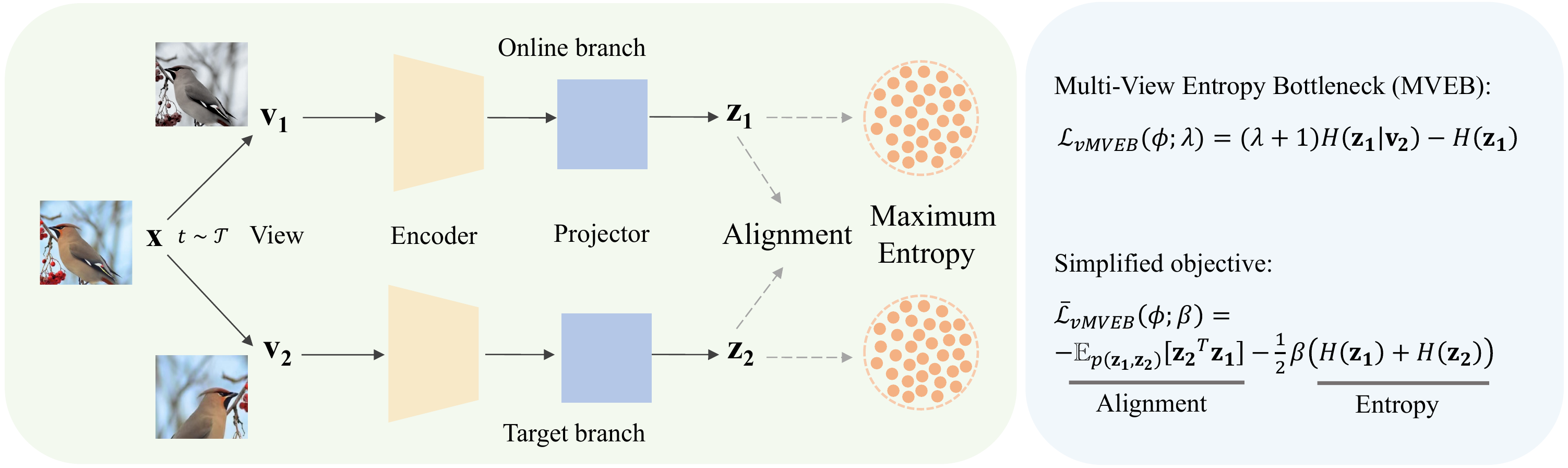}
\caption{Framework of MVEB and its training objective.}
\label{figure:MVEB}
\end{figure*}
Representation learning aims to transform input data $\mathbf{x}$ to the lower-dimensional representation $\mathbf{z}$ that contains the information relevant to the prediction task $\mathbf{y}$. 
This information is  considered  unchanged after encoding the data, which suggests 
$I(\mathbf{x} ; \mathbf{y})=I(\mathbf{z} ; \mathbf{y})$, where $I$ denotes mutual information. Thus, the learned representation $\mathbf{z}$ is sufficient for the prediction task~\cite{achille2018emergence}.

Since $\mathbf{x}$ has more information than $\mathbf{y}$, the sufficient representation $\mathbf{z}$ of $\mathbf{x}$ may contain superfluous information irrelevant to the prediction task. The superfluous information can be represented as conditional mutual information $I(\mathbf{x};\mathbf{z}|\mathbf{y})$. 
Among all sufficient representations, the minimal sufficient representation contains the least superfluous information.

%% file: doc/approach.tex
\vspace{1em}
\section{Approach}
We outline the general setting of training an encoder and a projector to learn a representation in the multi-view self-supervised setting in Fig.~\ref{figure:MVEB}. The Siamese
network consists of online  and target branches. Each branch includes an encoder and a projector. 
Let $\mathbf{v_1}$ and $\mathbf{v_2}$ be two different views of the input sample $\mathbf{x}$. 
We can get the $\ell_2$-normalized representations $\mathbf{z_1}$ and $\mathbf{z_2}$ from $\mathbf{v_1}$ and $\mathbf{v_2}$ through the deterministic function $f_{\phi}$ with the parameters $\phi$.  Let $q_{\phi}(\mathbf{z_1})$ and $q_{\phi}(\mathbf{z_2})$ be the marginal distributions of the representations $\mathbf{z_1}$ and  $\mathbf{z_2}$, respectively, which are used to compute the entropy $H(\mathbf{z_1})$ and $H(\mathbf{z_2})$.
  We derive a new objective $ \mathcal{L}_{vMVIB}(\phi;\lambda)$ to optimize the parameters $\phi$. 
  
\subsection{Multi-View Information Bottleneck}
\label{ss:MVIB}

\begin{figure}
\centering 
\includegraphics[width=0.4\textwidth]{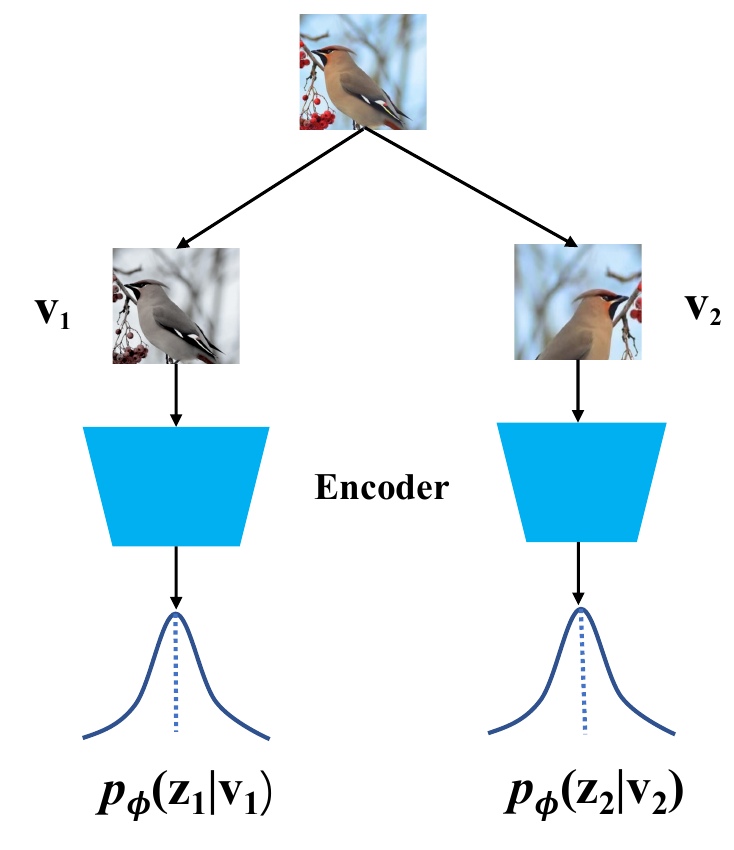}
\caption{Visualization of the multi-view information bottleneck model.}
\label{figure:MVIB}
\end{figure}

In the unsupervised setting, it is more challenging to
obtain the minimal sufficient representation since the superfluous information cannot be identified without downstream tasks. To overcome the problem,  \cite{DBLP:conf/iclr/Federici0FKA20} extends the information bottleneck theory in supervised learning to the multi-view unsupervised setting, termed  multi-view information bottleneck (MVIB).  
The main idea relies on the multi-view assumption that  either view  is (approximately) sufficient for the prediction of downstream tasks and contains the same task-relevant information.
In other words, the different views of a sample
should not affect the prediction for downstream tasks. 
Similar to the role of labels in supervised learning, two views can be considered as mutual self-supervised signals for each other. 
Hence, we can obtain the sufficiency for downstream tasks by ensuring that the representation $\mathbf{z_1} $ of $\mathbf{v_1}$ is sufficient for $\mathbf{v_2}$. The superfluous information can  also be identified as conditional mutual information $ I_\phi\left(\mathbf{z_1}; \mathbf{v_1} | \mathbf{v_2}\right)$. 
Decreasing $ I_\phi\left(\mathbf{z_1} ; \mathbf{v_1} | \mathbf{v_2}\right)$ can achieve the elimination of the superfluous information. 

We can learn the minimal sufficient representation by satisfying these requirements using the relaxed Lagrangian multiplier method:
\begin{align}\label{MIB}
     \mathcal{L}_{MVIB}\left(\phi ; \lambda\right)=\lambda I_\phi\left(\mathbf{z_1} ; \mathbf{v_1} | \mathbf{v_2}\right)- I_\phi\left(\mathbf{z_1} ; \mathbf{v_2}\right).
 \end{align}
Although MVIB is appealing in learning the minimal sufficient representation, the computation of the mutual information (MI) is notoriously intractable. 
To overcome the intractability  in MVIB, \cite{DBLP:conf/iclr/Federici0FKA20}  adopts the same stochastic network as VAE to obtain the Gaussian representation distributions $p(\mathbf{z_1}|\mathbf{v_1})$ and $p(\mathbf{z_2}|\mathbf{v_2})$ to approximately optimize 
$\mathcal{L}_{MVIB}$ as shown in Fig.~\ref{figure:MVIB}. 
Hence, we obtain the  variational MVIB objective as follows:

\begin{align}
   \mathcal{L}_{vMVIB}(\phi;\lambda) = &\lambda \wenlj{D_{KL}(p_{\phi}(\mathbf{z_1} |\mathbf{v_1} )||p_{\phi}(\mathbf{z_2} |\mathbf{v_2})) }\nonumber \\ & -I_\phi\left(\mathbf{z_1} ; \mathbf{z_2} \right),
\end{align}
where $D_{KL}$ denotes the Kullback–Leibler divergence. 
Specifically, \wenlj{$D_{KL}(p_{\phi}(\mathbf{z_1} |\mathbf{v_1} )||p_{\phi}(\mathbf{z_2} |\mathbf{v_2}))$} is the upper bound of  $I_\phi\left(\mathbf{z_1} ; \mathbf{v_1} | \mathbf{v_2}\right)$; $I_\phi\left(\mathbf{z_1} ; \mathbf{z_2} \right)$ is the lower bound of $I_\phi\left(\mathbf{z_1} ; \mathbf{v_2}\right)$ and  approximated by InfoNCE\cite{oord2018representation}, MINE\cite{belghazi2018mine} or MIGE\cite{DBLP:conf/iclr/WenZHZX20}.

\subsection{Multi-View Entropy Bottleneck}
\label{sec:MVEB}

\wenljx{MVIB cannot be directly applied to the Siamese network due to the intractability of the computation of mutual information. 
As mentioned in Section \ref{ss:MVIB}, the variational method can be introduced to optimize MVIB. However, 
this approximation optimization requires additional stochastic networks and does not work as expected in practice for visual recognition models compared with SimCLR, as shown in \cite{wang2022rethinking}. For the  Siamese network,  learning the minimal sufficient representation effectively in self-supervised representation learning is still
challenging.}

\wenljx{We derive the novel MVEB framework to solve the challenge of learning the minimal sufficient representation.} Compared with MVIB, MVEB can be directly applied to the Siamese network without modification of the network structure and other complex designs. 

The superfluous information $I_\phi\left(\mathbf{z_1} ; \mathbf{v_1} | \mathbf{v_2}\right)$ can  be decomposed \wenlj{(see Appendix)} as:
\begin{align}
\label{min_mi}
 I_\phi\left(\mathbf{z_1} ; \mathbf{v_1} |\mathbf{v_2}\right)=H(\mathbf{z_1}|\mathbf{v_2})-H(\mathbf{z_1}| \mathbf{v_1},\mathbf{v_2}),   
\end{align}
where the conditional entropy  $H(\mathbf{z_1}|\mathbf{v_1},\mathbf{v_2})$ contains no randomness (no information) as $\mathbf{z_1}$ being deterministic conditioned on $\mathbf{v}_1$. 
 Hence,  minimizing $H(\mathbf{z_1}|\mathbf{v_2})$  is equivalent to minimizing $I_\phi\left(\mathbf{z_1} ; \mathbf{v_1} |\mathbf{v_2}\right)$.
We can  also  decompose $I_\phi\left(\mathbf{z_1} ; \mathbf{v_2} \right)$ \wenlj{(see Appendix)} as:
\begin{align}
   \label{max_mi}
    I_\phi\left(\mathbf{z_1} ; \mathbf{v_2} \right)=H(\mathbf{z_1})-H(\mathbf{z_1}|\mathbf{v_2}).
\end{align}
Based on the above derivations and Eq.~(\ref{MIB}), we  obtain the general MVEB objective:
 \begin{align}
 \label{mvebsss}
      \mathcal{L}_{MVEB}\left(\phi ; \lambda\right)=(\lambda+1) H(\mathbf{z_1}|\mathbf{v_2})-H(\mathbf{z_1}),
 \end{align}
where $H(\mathbf{z_1}|\mathbf{v_2})=-\mathbb{E}_{ p(\mathbf{z_1}, \mathbf{v_2})} [\log p(\mathbf{z_1}|\mathbf{v_2})]$.
We can learn the minimal sufficient representation with $\mathcal{L}_{MVEB}$ for the deterministic encoder.

The conditional entropy $H(\mathbf{z_1}|\mathbf{v_2})$ is intractable since the distribution $p(\mathbf{z_1}|\mathbf{v_2})$ is unknown. To overcome this problem, we introduce $q_{\phi}(\mathbf{z_1}|\mathbf{v_2})$ that is a variational approximation to $p(\mathbf{z_1}|\mathbf{v_2})$. Since $KL(p(\mathbf{z_1}|\mathbf{v_2})||q_{\phi}(\mathbf{z_1}|\mathbf{v_2}))	\geq 0 $, we can derive the upper bound of $H(\mathbf{z_1}|\mathbf{v_2})$: 
\begin{align}\label{xxx}
H(\mathbf{z_1}|\mathbf{v_2})&=-\mathbb{E}_{ p(\mathbf{z_1}, \mathbf{v_2})} [\log p(\mathbf{z_1}|\mathbf{v_2})] \nonumber\\&\leq 
-\mathbb{E}_{ p(\mathbf{z_1}, \mathbf{v_2})} [\log q_{\phi}(\mathbf{z_1}|\mathbf{v_2})].
\end{align}
Hence, a variational MVEB term (vMVEB) is defined as 
 \begin{align}
 \label{vMVEB}
      \mathcal{L}_{vMVEB}\left(\phi ; \lambda\right)=&-(\lambda +1)\mathbb{E}_{ p(\mathbf{z_1}, \mathbf{v_2})}[\log q_{\phi}(\mathbf{z_1}|\mathbf{v_2})]\nonumber \\ &-H(\mathbf{z_1}). 
 \end{align}

\wenlj{For self-supervised learning based on the Siamese network, e.g. SimCLR, BYOL, and SimSiam,  the representation $\mathbf{z_1}$ is $\ell_2$-normalized in the hypersphere space to improve the performance of the models. We also normalize $\mathbf{z_1}$ in the hypersphere space. The von Mises-Fisher (vMF) is the common distribution of the hypersphere space. Hence,} we define $q_{\phi}(\mathbf{z_1}|\mathbf{v_2})$ as the vMF  distribution, i.e.,
\begin{align}
q_{\phi}(\mathbf{z_1}|\mathbf{v_2})=C_n(\kappa) e^{\kappa \boldsymbol{\mu}^T \mathbf{z_1}},
\end{align}
where $\boldsymbol{\mu}$ is the mean direction,  $\kappa$ denotes the concentration parameter of the von Mises-Fisher distribution and $C_n(\kappa)=\frac{\kappa^{\frac{n}{2} -1}}{(2 \pi)^{n / 2} I_{\frac{n}{ 2}-1}(\kappa)}$ is the normalization function of $\kappa$.
We assume that $\kappa$ is a constant and $q_{\phi}(\mathbf{z_1}|\mathbf{v_2})$ is  parameterized by $\boldsymbol{\mu}$. As illustrated in Fig.~\ref{figure:MVEB}, we use the target branch to encode $\mathbf{v_2}$ and output $\mathbf{z_2}$ as $\boldsymbol{\mu}$. 
Hence, we further obtain
\begin{align}
\label{centropy}
H(\mathbf{z_1}|\mathbf{v_2})\leq -\log C_n(\kappa)-\kappa \mathbb{E}_{ p(\mathbf{z_1}, \mathbf{z_2})} [\mathbf{z_2}^T\mathbf{z_1}],
\end{align} 
which  allows us to reformulate the objective of Eq.~(\ref{vMVEB}) as follows, \begin{align}\label{MVEB}
    \hat{\mathcal{L}}_{vMVEB}\left(\phi ; \beta\right)=-\mathbb{E}_{ p(\mathbf{z_1}, \mathbf{z_2})} [\mathbf{z_2}^T\mathbf{z_1}] -\beta H(\mathbf{z_1}),
\end{align}
where $\beta=\frac{1}{(\lambda+1)\kappa }$ is the  balance factor. 
This simplified objective maximizes both the agreement between $\mathbf{z_1}$ and $\mathbf{z_2}$ and the differential entropy of $\mathbf{z_1}$ to learn the minimal sufficient representation for the deterministic encoder.

The sample view  $\mathbf{v_1}$ can also be regarded as the 
self-supervised signal for $\mathbf{v_2}$. Similarly, we  derive  another optimization objective.  
The final simplified training objective is obtained as follows,
\begin{align}\label{MVEBf}
    \bar{\mathcal{L}}_{vMVEB}\left(\phi ; \beta\right)=&-\mathbb{E}_{ p(\mathbf{z_1}, \mathbf{z_2})} [\mathbf{z_2}^T\mathbf{z_1}] \nonumber\\ &- \frac{1}{2}\beta (H(\mathbf{z_1})+H(\mathbf{z_2})).
\end{align}

However, it is unfortunately intractable to compute $H(\mathbf{z_1})$  and $H(\mathbf{z_2})$  since the distributions of $\mathbf{z_1}$ and  $\mathbf{z_2}$  are unknown.
We propose a score-based entropy estimator with the von Mises-Fisher kernel to maximize $H(\mathbf{z_1})$ and  $H(\mathbf{z_2})$, which is described in Section~\ref{SEE}. The training pseudocode is given in  Algorithm~\ref{alg:method}.

\subsection{ Analysis of the Variational Approximation }
In our work, the variational approximation is used to obtain the upper bound of the conditional entropy $H(\mathbf{z_1}|\mathbf{v_2})$ for  minimization, rather than estimate the true $H(\mathbf{z_1}|\mathbf{v_2})$. However, we need to guarantee that this upper bound is not loose to achieve the minimization of $H(\mathbf{z_1}|\mathbf{v_2})$.

The  bound of the variational approximation of   $H(\mathbf{z_1}|\mathbf{v_2})$   in Eq.~(\ref{xxx}) is completely tight  if $\mathbb{E}_{ p(\mathbf{v_2})}[KL(p(\mathbf{z_1}|\mathbf{v_2})||q_{\phi}(\mathbf{z_1}|\mathbf{v_2}))]=0$. It means that $q_{\phi}(\mathbf{z_1}|\mathbf{v_2}))$ equals $p(\mathbf{z_1}|\mathbf{v_2})$. 
In other words, if $\mathbb{E}_{ p(\mathbf{v_2})}[KL(p(\mathbf{z_1}|\mathbf{v_2})||q_{\phi}(\mathbf{z_1}|\mathbf{v_2}))]$ is 
smaller, the  bound of the variational approximation is tighter. 
Below we prove the  bound of the variational approximation is not loose and can achieve the minimization of $H(\mathbf{z_1}|\mathbf{v_2})$.

If the approximation  is very loose, we cannot optimize $H(\mathbf{z_1}|\mathbf{v_2})=-\mathbb{E}_{ p(\mathbf{z_1}, \mathbf{v_2})} [\log p(\mathbf{z_1}|\mathbf{v_2})]$;
in other words, $-\mathbb{E}_{ p(\mathbf{z_1}, \mathbf{v_2})} [\log p(\mathbf{z_1}|\mathbf{v_2})]$  cannot be reduced during optimization. 
Decompose $\mathbb{E}_{ p(\mathbf{v_2})}[KL(p(\mathbf{z_1}|\mathbf{v_2})||q_{\phi}(\mathbf{z_1}|\mathbf{v_2}))]$ as follows:
\begin{align}
\label{xxxsss}
&\mathbb{E}_{ p(\mathbf{v_2})}[KL(p(\mathbf{z_1}|\mathbf{v_2})||q_{\phi}(\mathbf{z_1}|\mathbf{v_2}))]\nonumber\\ &=\mathbb{E}_{ p(\mathbf{z_1}, \mathbf{v_2})} [\log p(\mathbf{z_1}|\mathbf{v_2})] 
-\mathbb{E}_{ p(\mathbf{z_1},\mathbf{v_2})} [\log q_{\phi}(\mathbf{z_1}|\mathbf{v_2})].
\end{align}
Since  the first term  $\mathbb{E}_{ p(\mathbf{z_1}, \mathbf{v_2})} [\log p(\mathbf{z_1}|\mathbf{v_2})]$ of Eq.~(\ref{xxxsss}) is fixed, minimizing  $-\mathbb{E}_{ p(\mathbf{z_1}, \mathbf{v_2})} [\log q_{\phi}(\mathbf{z_1}|\mathbf{v_2})]$ is equivalent to minimizing $\mathbb{E}_{ p(\mathbf{v_2})}[KL(p(\mathbf{z_1}|\mathbf{v_2})||q_{\phi}(\mathbf{z_1}|\mathbf{v_2}))]$, which  makes the  bound  tight. 
\begin{algorithm}[H]
  \caption{  MVEB\ pytorch pseudocode.}
  \label{alg:method}
    \definecolor{codeblue}{rgb}{0.25,0.5,0.5}
    \definecolor{codekw}{rgb}{0.85, 0.18, 0.50}
    \newcommand{\algofontsize}{11.0pt}
    \lstset{
      backgroundcolor=\color{white},
      basicstyle=\fontsize{\algofontsize}{\algofontsize}\ttfamily\selectfont,
      columns=fullflexible,
      breaklines=true,
      captionpos=b,
      commentstyle=\fontsize{\algofontsize}{\algofontsize}\color{codeblue},
      keywordstyle=\fontsize{\algofontsize}{\algofontsize}\color{black},
    }
\begin{lstlisting}[language=python]
# f: encoder containing a backbone and a projector
# lambda:  loss balance coefficient
# N: batch size

for x in loader:  # load a minibatch x 

    # augmentation
    v_1, v_2 = augment(x)
    
    # compute  normalized embeddings
    z_1, z_2 = f(v_1), f(v_2)
    
    #  Alignment loss
    Align_loss = mm(z1, z2.t()).mean()
    
    # compute the score function S(.)
    # SGE: Stein Gradient Estimator
    S(z_1)= SGE(z_1)
    S(z_2)= SGE(z_2)

    # compute the entropy loss
    En_z_1= (S(z_1).detach()*z_1).sum(-1).mean()
    En_z_2= (S(z_2).detach()*z_2).sum(-1).mean()
    
    # compute the total loss
    loss = Align_loss+0.5*lambda*(En_z_1+ En_z_2)

    # optimization step
    loss.backward()
    optimizer.step()
\end{lstlisting}
\end{algorithm}
\subsection{Score-Based Entropy Estimation with the von Mises-Fisher Kernel}\label{SEE}
For learning the minimal sufficient representation with MVEB, we need to maximize $H(\mathbf{z})=-\mathbb{E}_{ q_{\phi}(\mathbf{z})} [\log q_{\phi}(\mathbf{z})]$. 
We first analyze the gradient of $H(\mathbf{z})$ w.r.t. $\phi$, which can be decomposed as:
\begin{align}\label{gentropy}
 \nabla_{\phi} H(\mathbf{z}) &= - \nabla_{\phi} \mathbb{E}_{ q_{\phi}(\mathbf{z})} [\log q(\mathbf{z})]- \mathbb{E}_{ q(\mathbf{z})} [\nabla_{\phi}\log q_{\phi}(\mathbf{z})], 
\end{align}
where $q(\mathbf{z})$ without the subscript $\phi$ means the gradient of computation is irrelevant to $\phi$.
The second term on the right part of Eq.~(\ref{gentropy}) can be further decomposed as:
\begin{align}
 \mathbb{E}_{ q(\mathbf{z})} [\nabla_{\phi}\log q_{\phi}(\mathbf{z})]
 = \mathbb{E}_{ q(\mathbf{z})} [\nabla_{\phi}q_{\phi}(\mathbf{z})\times \frac{1}{q(\mathbf{z})}] \nonumber\\=
 \int \nabla_{\phi}q_{\phi}(\mathbf{z})d \mathbf{z}= \nabla_{\phi}\int q_{\phi}(\mathbf{z})d \mathbf{z}=0.
\end{align}
Hence we have
\begin{align}
 \nabla_{\phi} H(\mathbf{z}) &=  -\nabla_{\phi} \mathbb{E}_{ q_{\phi}(\mathbf{z})} [\log q(\mathbf{z})].
\end{align}
However, $\nabla_{\phi} H(\mathbf{z})$ is non-trivial because the expectation  w.r.t. $q_\phi(\mathbf{z})$  is not differentiable  w.r.t. $\phi$. 

To overcome this problem, we adopt the general reparameterization trick proposed in \cite{roeder2017sticking} for the computation of $\nabla_{\phi} H(\mathbf{z})$.
In detail, the samples from the representation distribution $q_{\phi}(\mathbf{z})$ can be obtained by encoding the data samples $\mathbf{v}$, $\mathbf{z}=f_\phi(\mathbf{v})$, where  $f_\phi$ is the  deterministic function (encoder and projector).
Hence, the representation can be reparameterized via  the following differentiable transformation:
\begin{align}
 \mathbf{z}=f_{\phi}(\mathbf{v}) ~~~~~~\text{with} ~~~~~~\mathbf{v}~\sim  ~p(\mathbf{v}).
\end{align}
Since $p(\mathbf{v})$ is irrelevant to the model parameters $\phi$, the expectation w.r.t. $q_\phi(\mathbf{z})$ can be rewritten via the above reparameterization, which  makes the expectation differentiable w.r.t. $\phi$.
Hence, the entropy gradient estimator is derived as follows:
\begin{align}
\nabla_{\phi} H(\mathbf{z}) &=-\nabla_{\phi} \mathbb{E}_{ q_{\phi}(\mathbf{z})} [\log q(\mathbf{z})]=-\mathbb{E}_{ p(\mathbf{v})} [\nabla_{\phi}\log q(f_\phi (\mathbf{v}))]\nonumber\\&=-\mathbb{E}_{ p(\mathbf{v})} [\nabla_{\mathbf{z}}\log q(\mathbf{z}))\nabla_{\phi}f_{\phi}(\mathbf{v})],
\end{align}
where $\nabla_{\mathbf{z}}\log q(\mathbf{z})$ is the score function, which can be directly approximated by score estimation using a black-box function~\cite{DBLP:conf/iclr/LiT18}.  $\nabla_{\phi}f_{\phi}(\mathbf{x})$ can be obtained by direct back-propagation.
As long as we can provide a good enough approximation to the score function, this estimation of the entropy gradient is approximately unbiased.

The Stein gradient estimator  described in \wenlj{Appendix} is  an effective estimation of the score function~\cite{DBLP:conf/iclr/LiT18}.
We adopt it to approximate the score function $S(\mathbf{z})=\nabla_{\mathbf{z}}\log q(\mathbf{z})$. 
Since the representation $\mathbf{z}$ is $\ell_2$-normalized,  we propose to
 use the following von Mises-Fisher kernel to compute  \wenljx{$S(\mathbf{z})$ }:
\begin{align}
\label{Gram}
k\left(\mathbf{z}, \mathbf{z}^{\prime}\right)=\exp \left(\frac{\mathbf{z}^T\mathbf{z}^{\prime}}{ \triangle }\right),
\end{align}
where $\triangle $ is the bandwidth of the von Mises-Fisher kernel. We set it to the median of pairwise cosine  distances among all samples in the batch. 


\section{Rethinking Alignment and Uniformity}
\label{RAU}

Contrastive learning (e.g., SimCLR and MOCO) aims to bring similar (positive) samples closer and dissimilar (negative) samples farther apart. \cite{DBLP:conf/icml/0001I20} decomposes the contrastive loss  into alignment and uniformity. As shown in \cite{tao2022exploring}, asymmetric network methods (e.g., BYOL and SimSiam) and feature decorrelation methods (e.g., Barlow Twins and  VICReg) are viewed as optimizing alignment and uniformity based on gradient analysis.  
Asymmetric network methods rely on a predictor to optimize the uniformity,  and feature decorrelation methods rely on feature decorrelation to optimize the uniformity.

 We consider the multi-view self-supervised setting in Fig. \ref{figure:MVEB}, where $\mathbf{z_1}$ and $\mathbf{z_2}$ are the representations of views $\mathbf{v_1}$ and  $\mathbf{v_2}$, respectively. 
 If we consider  $\mathbf{v_2}$ as the supervised information,  minimizing superfluous information $I_\phi\left(\mathbf{z_1} ; \mathbf{v_1} |\mathbf{v_2}\right)$ 
 is equivalent to  minimizing $H(\mathbf{z_1}|\mathbf{v_2})$   for the deterministic encoder, as shown in Section~\ref{sec:MVEB}. Furthermore, maximizing  the alignment $\mathbb{E}_{ p(\mathbf{z_1}, \mathbf{z_2})} [\mathbf{z_2}^T\mathbf{z_1}]$ is equivalent to  minimizing  $H(\mathbf{z_1}|\mathbf{v_2})$ (see Eq.~(\ref{centropy})).
 Hence, maximizing  the alignment can  eliminate superfluous information. 

We find that maximizing the mutual information $I_\phi(\mathbf{z_1}; \mathbf{v_2})$ between  $\mathbf{z_1}$ and the supervised information $\mathbf{v_2}$ not only keeps the information relevant to $\mathbf{v_2}$ but also reduces superfluous information. 
This is because  $ I_\phi\left(\mathbf{z_1} ; \mathbf{v_2} \right)=H(\mathbf{z_1})-H(\mathbf{z_1}|\mathbf{v_2})$ and minimizing superfluous information $I_\phi\left(\mathbf{z_1} ; \mathbf{v_1} |\mathbf{v_2}\right)$ 
 is equivalent to  minimizing $H(\mathbf{z_1}|\mathbf{v_2})$   for the deterministic encoder.
As uniformity prefers the feature distribution $p(\mathbf{z_1})$ that preserves its maximal entropy $H(\mathbf{z_1})$,  we can consider $H(\mathbf{z_1})$ as the uniformity. Hence, $I_\phi(\mathbf{z_1}; \mathbf{v_2})$  is the combination of alignment and uniformity.
In another view, maximizing  alignment and uniformity can keep the information in $\mathbf{z_1}$
relevant to $\mathbf{v_2}$.  
However, since the superfluous information is not minimal, maximizing $I_\phi(\mathbf{z_1}; \mathbf{v_2})$ cannot achieve the goal of learning of minimal sufficient representation.

In this work, we present a new objective
function, $\mathcal{L}_{MVEB}\left(\phi; \lambda\right)=(\lambda+1) H(\mathbf{z_1}|\mathbf{v_2})-H(\mathbf{z_1})$, to
learn the minimal sufficient representation. $H(\mathbf{z_1}|\mathbf{v_2})-H(\mathbf{z_1})= -I_\phi(\mathbf{z_1}; \mathbf{v_2})$ aims to  keep the information relevant to $\mathbf{v_2}$, and $\lambda H(\mathbf{z_1}|\mathbf{v_2})$ aims to reduce superfluous information $I_\phi\left(\mathbf{z_1} ; \mathbf{v_1} |\mathbf{v_2}\right)$.
Since  maximizing  the alignment $\mathbb{E}_{ p(\mathbf{z_1}, \mathbf{z_2})} [\mathbf{z_2}^T\mathbf{z_1}]$ is equivalent to  minimizing  $H(\mathbf{z_1}|\mathbf{v_2})$,  the balance of  maximizing alignment and uniformity can learn  minimal sufficient representation.
Specifically, the coefficient $\beta$ in Eq.~(\ref{MVEBf}) is used to balance the optimization of the alignment and  $H(\mathbf{z_1})+H(\mathbf{z_1})$. In Eq.~(\ref{MVEB}), $\beta=\frac{1}{(\lambda+1)\kappa }$, where $\kappa$ is a  constant. As shown in Eq.~(\ref{MIB}), $\lambda$ is the coefficient to balance the optimization of $I_\phi\left(\mathbf{z_1} ; \mathbf{v_1} | \mathbf{v_2}\right)$ and $I_\phi\left(\mathbf{z_1} ; \mathbf{v_2}\right)$. 
Increasing $\beta$
to make $\lambda$  smaller than a threshold does not  eliminate superfluous information effectively, which hurts the  performance of downstream tasks.
Decreasing $\beta$ to  approach zero means  $\lambda$ approaches infinity, and the Siamese network suffers from model collapse with the trivial constant representations without maximizing the uniformity. 
In anther view, when $\lambda$ approaches infinity, the optimization objective in Eq.~(\ref{MIB}) only considers   minimizing superfluous information $I_\phi\left(\mathbf{z_1} ; \mathbf{v_1} |\mathbf{v_2}\right)$.
Hence, a trivial solution is obtained if constant representations of  $\mathbf{v_1}$ are outputted, which means the representations do not contain  the information relevant to $\mathbf{v_1}$ and  $\mathbf{v_2}$.

\vspace{0.5em}
\noindent\textbf{Relation with Contrastive Learning.}
The contrastive loss, also termed InfoNCE, is a lower bound of  $I_\phi(\mathbf{z_1}; \mathbf{v_2})$~\cite{DBLP:conf/iclr/Tsai0SM21}.
More negative samples make this lower bound tighter. 
Maximizing the contrastive loss aims to maximize $I_\phi(\mathbf{z_1}; \mathbf{v_2})$. Specifically, when the number of negative samples $N \rightarrow \infty$, the normalized contrastive loss reaches the following convergence:
\begin{align}
\label{contrastive}
\lim _{N \rightarrow \infty} &\mathcal{L}_{\text {contrastive }}(\phi ; \tau, N)-\log N\nonumber\\
 = &-\frac{1}{\tau} \underset{(\mathbf{v_1}, \mathbf{v_2}) \sim p_{\text {pos }}}{\mathbb{E}}\left[f_\phi(\mathbf{v_1})^{\top} f_\phi(\mathbf{v_2})\right] \nonumber\\
& +\underset{\mathbf{v_1}\sim p_{\text {data }}}{\mathbb{E}}\left[\log \underset{\mathbf{v_1^{-}} \sim p_{\text {data } } \backslash \mathbf{v_1}} {\mathbb{E}}\left[e^{f_\phi\left(\mathbf{v_1^{-}}\right)^{\top} f_\phi(\mathbf{v_1}) / \tau}\right]\right],
\end{align}
where  $\mathbf{v_1^{-}}$ denotes the negative samples of $\mathbf{v_1}$ and $p_{\text {pos}}$ denotes the distribution of the pairs of positive samples. 
The first term of the right-hand side of Eq.~(\ref{contrastive}) aims to maximize alignment. 
The second term  pushes dissimilar (negative)
samples farther apart.  
The performances of contrastive learning are sensitive to the choice of the hyper-parameter $\tau$, since $\tau$ is used to  balance the optimization of alignment and uniformity to learn minimal sufficient representation. 
However, $\tau$ is also incorporated to maximize uniformity, which limits the balance between alignment and uniformity based on the gradient analysis in \cite{tao2022exploring}.  
 
Contrastive learning relies on a large number of negative samples to optimize uniformity and keep a lower bound of $I(\mathbf{z_1}; \mathbf{v_2})$ tight.  As uniformity optimization is based on the separation of the instances,  it is hard to maximize global uniformity effectively. 
Unlike contrastive learning,   our MVEB directly maximizes the differential entropy of the global feature distribution, which is, in principle, more effective for uniformity maximization.

\vspace{0.5em}
\noindent\textbf{Relation with Asymmetric Network Methods  and Feature
Decorrelation Methods. }
As shown in \cite{tao2022exploring}, from the gradient analysis, asymmetric network methods  and feature decorrelation methods can be unified into the same  form:
\begin{align}
\label{sssssddsss}
 \mathcal{L}\left(\phi ; \lambda\right)=-\mathbb{E}_{ p(\mathbf{z_1}, \mathbf{z_2})} [\mathbf{z_2}^T\mathbf{z_1}] +\lambda\mathbb{E}_{ p(\mathbf{z_1})}[\mathbf{z_1}^TF\mathbf{z_1}],
\end{align}
where $F$ is  the correlation matrix of features; $\lambda$ is the banlance factor.  
For asymmetric network methods, $F$ is updated according to the moving average; for feature decorrelation methods, $F$ is computed according to the features of each batch.  
The first term of the right-hand side of Eq.~(\ref{sssssddsss}) is the alignment. 
The second term  is the derivation of the following entry:
\begin{align}
\mathbb{E}_{p(\mathbf{z_1^-})}[\cos ^2\left(\mathbf{z_1}, \mathbf{z_1^-}\right)]=\mathbb{E}_{p(\mathbf{z_1^-})}[\mathbf{z_1}^T\mathbf{z_1^-}\mathbf{z_1^-}^T\mathbf{z_1}]\nonumber\\=\mathbb{E}_{p(\mathbf{z_1^-})}[\mathbf{z_1}^T(\mathbf{z_1^-}\mathbf{z_1^-}^T)\mathbf{z_1}]=\mathbf{z_1}^TF\mathbf{z_1},
\end{align}
where $\mathbf{z_1^-}$ denotes the negative samples of $\mathbf{z_1}$. According to this derivation, the second term of Eq.~(\ref{sssssddsss}) aims to minimize the similarity between negative samples to maximize uniformity.
Hence,  asymmetric network methods  and feature decorrelation methods also achieve balancing the optimization of alignment and uniformity to learn minimal sufficient
representation. 

Asymmetric network methods rely on a predictor to optimize uniformity; feature decorrelation methods rely on feature decorrelation to optimize uniformity.   
This difference leads to the difference in performances.
Although the performances of these methods are better than contrastive learning, they still rely on separating the instances, which poses difficulty in achieving optimal results.
We argue that maximizing the differential entropy is the better way to learn the minimal sufficient
representation with the feature distribution rather than the instance separation.


%% file: doc/exp.tex
\section{Main Results}
We first assess MVEB's representation with the self-supervised benchmark on ImageNet dataset~\cite{deng2009imagenet} in linear evaluation. We then evaluate our model by transferring it to other datasets and tasks, including image classification, object detection, and segmentation.

\subsection{Pretraining  Details}
\label{sspdss}
We pretrain our model MVEB on  ImageNet with ResNet-50 as the backbone. The projector network consists of three linear layers,  each with an output dimensionality set to 2048. We apply BN and ReLU after the first two layers. According to our empirical study, the momentum encoder is chosen as the target branch (see Fig.~\ref{figure:MVEB}). Following the setting of BYOL~\cite{DBLP:conf/nips/GrillSATRBDPGAP20}, we update the target branch with the exponential moving by increasing the average parameters from 0.996 to 1 with a cosine scheduler. 

We follow the strategy of image augmentation used in BYOL, including random cropping, color jittering, converting to grayscale,  horizontal flipping,
Gaussian blurring and solarization. We also adopt multi-crop to get six local views~\cite{caron2021emerging} of $96\times96$. All augmentation parameters are the same as those in DINO~\cite{caron2021emerging}. The local views are only passed through the online branch. In addition, the positive
sample of each local view only comes from the average of the embeddings of the two global views~\cite{caron2021emerging} from the same sample.

We train MVEB for 800 epochs with the LARS~\cite{you2017scaling} optimizer.  The weight decay and the momentum are set to 1e-6 and 0.9, respectively.  The basic learning rate is 0.4, scaled with the batch size and divided by 256.
We decrease the learning rate
to one-thousandth with a cosine decay scheduler after a warm-up period of 10 epochs. The biases and
batch normalization parameters are excluded from the LARS adaptation. The batch size is 4096, distributed over 32 NVIDIA V100 GPUs. The coefficient $\beta$ in the loss function is set to 0.01  according to our empirical study.

\begin{table*}[!t]
	\caption{Top-1 and top-5 accuracies (\%) of linear classification on ImageNet~\cite{deng2009imagenet}. The bold entries denote the best. The results of all methods are based on the ResNet-50~\cite{he2016deep} backbone for a fair comparison.}	
	\begin{center}
		\tablestyle{9pt}{1.2}
		\begin{adjustbox}{max width=1\textwidth}
		\begin{tabular}{lccccc}
			\shline
			Method               &  Batch size   &                Training Epoch & Multi-crop   & Top-1            & Top-5 \\
			\hline

		SimCLR \cite{DBLP:conf/icml/ChenK0H20}     &  4096  &  1000   & 2$\times$224            & 69.3 &   89.0              \\

			MoCo-v2 \cite{chen2020improved}  & 256 &  800  &  2$\times$224            & 71.1 & 90.1      \\
InfoMin~\cite{tian2020makes}  & 256 &  800  &  2$\times$224 &          73.0 &  91.1\\
   
   	 SimSiam~\cite{chen2021exploring}   & 256 &  800  &  2$\times$224 &           71.3 & -   
 
			\\
          Barlow Twins~\cite{zbontar2021barlow}                                 &   2048  &1000 &   2$\times$224  &73.2 &{91.0} \\
 VICReg~\cite{DBLP:conf/iclr/BardesPL22}                           &   2048  &1000 &   2$\times$224   & 73.2 &{91.1} \\

                BYOL \cite{DBLP:conf/nips/GrillSATRBDPGAP20}                                              &  4096  & 1000  &  2$\times$224   &74.3 & 91.6\\
               
 MVEB (ours) & 4096       &   800      &    2$\times$224      & \textbf{74.6} & \textbf{92.1}\\	
 \hline
			SwAV~\cite{DBLP:conf/nips/CaronMMGBJ20}                                        & 4096       &   800      &    2$\times$224 + 6$\times$96    & 75.3& - \\
		Self-Classifier~\cite{amrani2021self}	& 4096       &   800      &    2$\times$224 + 6$\times$96    & 74.1& - \\
	DINO~\cite{caron2021emerging}	& 4096       &   800      &    2$\times$224 + 8$\times$96    & 75.3 & 92.5	\\
    UniGrad~\cite{tao2022exploring} & 4096       &   800      &    2$\times$224 + 6$\times$96   & 75.5 & -\\

      MVEB (ours) & 4096       &   800      &    2$\times$224 + 6$\times$96     & \textbf{76.9} & \textbf{93.3}\\	
			
   \shline
		\end{tabular}
		\end{adjustbox}
	\end{center}
	\label{tab:imagenet_main}
\end{table*}

\subsection{Linear Evaluation on ImageNet}
\label{subsec:lineareval}

\begin{table}
\caption{\wenlj{Comparison of MVEB with masked autoencoder-based methods for linear evaluation on ImageNet~\cite{deng2009imagenet}. The bold entry denotes the best.} }	
\begin{center}
\tablestyle{9pt}{1.2}
\begin{adjustbox}{max width=0.48\textwidth}
\begin{tabular}{lccccc}

\shline
Method  &  Batch size   &  Epoch & Backbone  & Param. & Top-1 \\
\hline
MAE\cite{DBLP:conf/cvpr/HeCXLDG22}	& 4096	& 1600 & ViT-B &	85M & 67.8 \\
MAE\cite{DBLP:conf/cvpr/HeCXLDG22}	& 4096	& 1600 & ViT-L &	307M & 76.0\\	
SimMM\cite{DBLP:conf/cvpr/Xie00LBYD022}& 2048 & 800& ViT-B &	85M &56.7 \\
\hline
 MVEB (ours)& 4096	& 800 & ResNet-18 &	11.7M  &	64.9\\
     MVEB (ours)& 4096	& 800 & ResNet-34 &	21.8M  &	69.6\\
     MVEB (ours)& 4096	& 800 & ResNet-50 &	23M  &	\textbf{76.9}\\
        MVEB (ours)& 4096	& 800 & ResNet-101 &		44.5M  &	\textbf{78.2}
\\			
   \shline
	\end{tabular}
	\end{adjustbox}
	\end{center}
	\label{tab:mask}
\end{table}

Following the ImageNet linear evaluation  in~\cite{DBLP:conf/eccv/ZhangIE16, DBLP:conf/icml/ChenK0H20, DBLP:conf/nips/GrillSATRBDPGAP20} and \cite{DBLP:conf/cvpr/He0WXG20}, we train a linear classifier on top of the frozen learned representation to assess the classification performance on ImageNet~\cite{deng2009imagenet}. 
The number of training epochs  is set to 50. Other training  settings of the linear evaluation are kept the same as~\cite{chen2021exploring}. 

We compare MVEB with previous popular SSL methods \wenlj{based on the Siamese network}. The results are shown in Table \ref{tab:imagenet_main}. MVEB significantly exceeds the previous best  method UniGrad in top-1  accuracy by an absolute improvement of 1.4\%. Compared with the supervised baseline used in \cite{ DBLP:conf/icml/ChenK0H20}, MVEB surpasses the supervised result of 76.5\%.
To the best of our knowledge, MVEB is the first work
that exceeds this supervised  learning result with the vanilla ResNet-50 backbone.

\wenlj{We compare MVEB with  the masked autoencoder-based methods, MAE \cite{DBLP:conf/cvpr/HeCXLDG22} and  SimMM \cite{DBLP:conf/cvpr/Xie00LBYD022}. Table \ref{tab:mask} shows the result. For linear evaluation on ImageNet, MVEB outperforms MAE and SimMM. 
Moreover, the number of parameters of MVEB is the smallest.}


\subsection{Semi-Supervised Classification on ImageNet}
We implement the semi-supervised learning  by fine-tuning the pre-trained MVEB on both the 1\% and 10\% subsets of the ImageNet training set, utilizing the same partitions as in SimCLR. Adhering to the semi-supervised training configurations outlined in~\cite{DBLP:conf/iclr/BardesPL22}, we  train a linear classifier and fine-tune the representations using 1\% and 10\% of the available labels.
Our training employs the SGD optimizer with no weight decay, a batch size of 256, and running for 60 epochs. 
For the training with  1\% of labels, we use a learning rate of 0.002 for the encoder and a learning rate of 0.8  for the linear head.
For the training with  10\% of labels, we use a learning rate of 0.003 for the encoder  and a learning rate of 0.4 for the linear head.
The cosine decay is employed to adjust the two learning rates.
The augmentation pipelines used for training data and validation are the same as those  for augmenting the data in linear evaluation.

In Table~\ref{table:SS_C}, we present the top-1 and top-5 accuracies. 
Our results indicate that MVEB outperforms previous methods consistently in both the 1\% and 10\% settings. Additionally, it is worth mentioning that all self-supervised learning methods significantly outperform the supervised baseline\cite{9010283}.

\begin{table}
\caption{Semi-supervised classification results on ImageNet. We use
 1\% and  10\% training examples to fine-tune the pre-trained model.   }
\begin{center}
\tablestyle{9pt}{1.2}
\begin{adjustbox}{max width=1\textwidth}
\begin{tabular}{lcccc}
\shline
\multirow{2}{*}{ Method } & \multicolumn{2}{c}{$1 \%$} & \multicolumn{2}{c}{$10 \%$} \\
& Top 1 & Top 5 & Top 1 & Top 5 \\
\hline 
Supervised\cite{9010283} & 25.4 & 48.4 & 56.4 & 80.4 \\
\hline 
SimCLR\cite{DBLP:conf/icml/ChenK0H20}  & 48.3 & 75.5 & 65.6 & 87.8 \\
BYOL \cite{DBLP:conf/nips/GrillSATRBDPGAP20} & 53.2 & 78.4 & 68.8 & 89.0 \\
Barlow Twins \cite{zbontar2021barlow} & 55.0 & 79.2 & 69.7 & 89.3 \\
DINO \cite{caron2021emerging} & 52.2 & 78.2 & 68.2 & 89.1 \\
VICReg~\cite{DBLP:conf/iclr/BardesPL22} & 54.8 & 79.4 & 69.5 & 89.5 \\
\hline 
MVEB (ours) & \bf 57.5 &\bf 82.1 & \bf 72.6 & \bf 91.5 \\
\shline
\end{tabular}
\end{adjustbox}
\end{center}
\label{table:SS_C}
\end{table}

\subsection{Transfer Learning}
To assess whether our learned representation is generic across different domains, we transfer it to other classification tasks on 11 datasets, including FGVC-Aircraft~\cite{maji2013fine}, Caltech-101~\cite{fei2004learning}, Stanford Cars~\cite{krause2013collecting}, CIFAR-10~\cite{krizhevsky2009learning}, CIFAR-100~\cite{krizhevsky2009learning}, Describable Textures Dataset (DTD)~\cite{cimpoi2014describing}, Oxford 102 Flowers~\cite{nilsback2008automated}, Food-101~\cite{bossard2014food}, Oxford-IIIT Pets~\cite{parkhi2012cats}, SUN397~\cite{xiao2010sun}
and Pascal VOC2007~\cite{everingham2010pascal}. 
For each dataset, we conduct (a) linear evaluation, where a multinomial logistic regression model is fit on top of the embeddings extracted from the frozen ResNet-50 backbone, and (b) fine-tuning, where the weights of both the backbone and classifier are allowed to be updated. We search for the best hyperparameters ($\ell_2$-regularization coefficient for linear evaluation and learning rate and weight decay for fine-tuning) on the split validation set and report the evaluation reuslt on the test set of each dataset. 

Following the common practice in~\cite{DBLP:conf/icml/ChenK0H20} and \cite{DBLP:conf/nips/GrillSATRBDPGAP20}, we evaluate the transfer performance on the 11 datasets via linear classification and fine-tuning. As for the evaluation, we use the metrics in the papers introducing these datasets. Specifically, we report top-1 accuracy for CIFAR-10, CIFAR-100, DTD, Food-101, Stanford Cars and SUN397, mean per-class accuracy for Caltech-101, FGVC-Aircraft, Oxford-IIIT Pets, and Oxford 102 Flowers, and 11-point mAP  from ~\cite{everingham2010pascal} for Pascal VOC 2007. For DTD and SUN397, which contain multiple train/test splits defined by the original creators, we only report results in the first train/test split. For Caltech-101, since there is no defined train/test split, we randomly select 30 images per class to form the training set and the rest is for test.  DTD, FGVC-Aircraft,  Pascal VOC 2007, and Oxford 102 Flowers have their own validation sets, and we directly use them. For the others, we hold out a subset by randomly selecting 20\% from the training set to form the validation set. The hyperparameters are chosen based on the metrics on the split validation set, and the final results are reported on the test set.

\vspace{0.5em}
\noindent\textbf{Linear Classification.} We fit an $\ell_2$-regularization multinomial logistic regression model on top of the  embeddings extracted from the frozen ResNet-50 backbone. The images are resized to 224 pixels along the shorter side using bicubic resampling, after which  a center crop of $224\times224$ is followed. We use L-BFGS\cite{XIAO20081001} to optimize the softmax cross-entropy objective. The coefficient of the $\ell_2$-regularization for each dataset is chosen on the validation set, ranging over a grid of 45 logarithmically spaced values between $10^{-6}$ and $10^{5}$. 

\begin{table*}
\caption{Transfer learning results of the models pretrained on ImageNet. The best result for each dataset is \textbf{bold}, and the second best is \underline{underlined}. `Supervised' refers to the pretrained model provided in the PyTorch library~\cite{paszke2019pytorch}, where the model is pretrained using the labels of ImageNet. The results of all methods are based on the vanilla ResNet-50 backbone.}
\begin{center}

\begin{adjustbox}{max width=1\textwidth}
\begin{tabular}{l c c c c c c c c c c c |c}
\cmidrule[\heavyrulewidth]{1-13}
{ Method} & { Aircraft} & { Caltech101} & { Cars} & { CIFAR10} & { CIFAR100}  & {DTD} & { Flowers} & { Food} & { Pets} & { SUN397} & { VOC2007} & { Avg.} \\
\midrule
\multicolumn{13}{l}{\emph{Linear evaluation}}\\
\midrule
Supervised  &              43.6 &              90.2 &              44.9 &              91.4 &              73.9 &              72.2 &              89.9 &              69.5 &     \bf{91.5} &              60.5 &              \underline{83.6} &              73.8 \\
InfoMin~\cite{tian2020makes}&              38.6 &              87.8 &              41.0 &              91.5 &              73.4 &              74.7 &              87.2 &              69.5 &              86.2 &              61.0 &              83.2 &              72.2 \\
SimCLR~\cite{DBLP:conf/icml/ChenK0H20} & 44.9 & 90.1 & 43.7 & 91.2 & 72.7 & 74.2 & 90.9 & 67.5 & 83.3 & 59.2 & 80.8 & 72.6 \\
MoCo v2~\cite{chen2020improved} &              41.8 &              87.9 &              39.3 &              92.3 &              74.9 &              73.9 &              90.1 &              69.0 &              83.3 &              60.3 &              82.7 &              72.3 \\
SimCLR v2~\cite{chen2020big}&              46.4 &              89.6 &              50.4 &              92.5 &              76.8 &              76.4 &              92.9 &              \underline{73.1} &              84.7 &              \underline{61.5} &              81.6 &              75.1 \\
BYOL~\cite{DBLP:conf/nips/GrillSATRBDPGAP20} &              \underline{53.9} &    \underline{91.5} &  \underline{56.4} &              \underline{93.3} &              \underline{77.9} &              \bf{76.9} &              \underline{94.5} &              73.0 &              89.1 &              60.0 &              81.1 &              \underline{77.1} \\
\midrule
MVEB (ours)& \bf{55.8} & \bf{92.7} & \bf{59.6} & \bf{94.6} & \bf{79.4} & \underline{76.5} & \bf{95.7} & \bf{77.9} & \underline{91.2} & \bf{66.5} & \bf{85.5} & \bf{79.6} \\ 
\midrule
\multicolumn{13}{l}{\emph{Fine-tuning}}\\
\midrule
Supervised     &     \underline{83.5} &     \bf{91.0} &              82.6 &              96.4 &              82.9 &              73.3 &  \underline{95.5} &              84.6 &   \bf{92.4} &              63.6 &              \underline{84.8} &              \underline{84.6} \\
InfoMin~\cite{tian2020makes}& 80.2 &              83.9 &              78.8 &              96.9 &              71.2 &              71.1 &              95.2 &              78.9 &              85.3 &              57.7 &              76.6 &              79.6 \\
SimCLR~\cite{DBLP:conf/icml/ChenK0H20}&              81.1 &              \underline{90.4} &              83.8 &     \bf{97.1}  & \bf{84.5} &              71.5 &              93.8 &              82.4 &              84.1 &              63.3 &              82.6 &              83.1 \\
MoCo v2~\cite{chen2020improved}  &              79.9 &              84.4 &              75.2 &              96.5 &              71.3 &              69.5 &              94.4 &              76.8 &              79.8 &              55.8 &              71.7 &              77.7 \\
SimCLR v2~\cite{chen2020big}&              78.7 &              82.9 &              79.8 &              96.2 &              79.1 &              70.2 &              94.3 &              82.2 &              83.2 &              61.1 &              78.2 &              80.5 \\
BYOL~\cite{DBLP:conf/nips/GrillSATRBDPGAP20} &              79.5 &              89.4 &              \underline{84.6} &              \underline{97.0} &               \underline{84.0} &              \underline{73.6} &              94.5 &              \bf{85.5} &  89.6 &              \underline{64.0} &              82.7 &              84.0 \\
\midrule
MVEB (ours) & \bf{85.9} & 88.6 & \bf{89.9} & \underline{97.0} & 80.4 & \bf{74.5} & \bf{95.7} & \underline{84.9} & \underline{91.9} & \bf{64.3} & \bf{85.6} & \bf{85.3}  \\
\bottomrule

\end{tabular}
\end{adjustbox}
\end{center}

\label{table:transfer}
\end{table*}
\vspace{0.5em}
\noindent\textbf{Fine-Tuning.} We initialize the model with the parameters of the pretrained model and tune the whole network. For augmentation, we only perform random crops with resizing and flipping at training time. With a batch size of 64, we train the model for 5000 iterations. The optimizer is SGD with the Nesterov momentum with the parameter set to 0.9. The learning rate is decreased with the cosine annealing schedule without restart.  We search for the best learning rate and weight decay on the validation set. Specifically, the initial learning rate is chosen from  a grid
of 4 logarithmically spaced values between 0.0001 and 0.1, and the weight decay is chosen from a grid of 4 logarithmically spaced values between $10^{-6}$
and $10^{-3}$, as well as no weight decay. The values of weight decay are divided by the learning rates.

As shown in Table \ref{table:transfer},  for linear evaluation, MVEB outperforms other methods by a large margin on all datasets except for DTD and Pets where the results of MVEB are still competitive. In the case of fine-tuning, MVEB also achieves the best or second best on 9 of 11 datasets, surpassing the supervised baseline in terms of the average evaluation metric of all datasets. Compared with other SSL methods, MVEB shows more advantages in generalization across different image domains.

\subsection{Object Detection and Segmentation}
\begin{table*}[h]
 \caption{Object detection and instance segmentation results (\%) on MS COCO. }
 \begin{center}
\tablestyle{9pt}{1.2}
\begin{adjustbox}{max width=1\textwidth}
    \begin{tabular}{@{}cccc|ccc@{}}
    \toprule
    \multirow{2}{*}{Method} & \multicolumn{3}{c}{COCO object detection} & \multicolumn{3}{c}{COCO instance segmentation} \\
    & AP$_\mathrm{all}^{\mathrm{box}}$ & AP$_{50}^{\mathrm{box}}$ & AP$_{75}^{\mathrm{box}}$ & AP$_\mathrm{all}^{\mathrm{mask}}$ & AP$_{50}^{\mathrm{mask}}$ & AP$_{75}^{\mathrm{mask}}$\\
    \hline
    Supervised  & 38.9 & 59.6 & 42.7 & 35.4 & 56.5 & 38.1 \\
    \hline
    MoCo v2~\cite{chen2020improved} & 39.8 & 59.8 & 43.6 & 36.1 & 56.9 & 38.7  \\
    DenseCL~\cite{wang2021dense} & 40.3 &59.9 & 44.3 & 36.4 & 57.0 & 39.2 \\
    DC v2~\cite{DBLP:conf/nips/CaronMMGBJ20} & 41.0 & 61.8 & 45.1 & 37.3 & 58.7 & 39.9 \\
    DINO~\cite{caron2021emerging} & 41.4 & 62.2 & 45.3 & 37.5 & 58.8 & 40.2  \\
    SimCLR~\cite{DBLP:conf/icml/ChenK0H20} & 41.6 & 61.8 & 45.6 & 37.6 & 59.0 & 40.5 \\ 
    UniGrad~\cite{tao2022exploring} & 42.0 & 62.6 & 45.7 & 37.9 & 59.7 & 40.7 \\
    \hline
     MVEB (ours) & \textbf{42.2} & \textbf{62.8} & \textbf{46.2} & \textbf{38.1} & \textbf{59.8} & \textbf{41.1} \\ 
    \bottomrule
    \end{tabular}
    \end{adjustbox}
	\end{center}
    \label{tab:coco}

\end{table*}
We further evaluate the learned embeddings  by transferring them to more downstream tasks besides  classification, including 
object detection and instance segmentation on MS COCO~\cite{DBLP:conf/eccv/LinMBHPRDZ14}. We adopt ResNet-50~\cite{he2016deep} with the  feature pyramid network (FPN)~\cite{DBLP:conf/eccv/LinMBHPRDZ14}  and 
Mask RCNN~\cite{he2017mask} for detection and segmentation. The ResNet-50 backbone is pretrained by MVEB for 800 epochs, as in Section \ref{subsec:lineareval}. For implementation, we adopt Detectron2~\cite{wu2019detectron2} and use the hyperparameters suggested in \cite{he2020momentum} without searching for the best hyperparameters. The model is fine-tuned on  COCO 2017 with the  $1\times$ training schedule\cite{DBLP:conf/cvpr/He0WXG20}.

The results are reported in Table \ref{tab:coco}. It is shown that our MVEB consistently outperforms other methods on both object detection and instance segmentation  w.r.t. all evaluation metrics. This indicates that MVEB's representation generalizes well beyond the ImageNet classification task.

\section{Empirical Study}
In this section, we explore the behaviors of MVEB in self-supervised learning with Siamese networks.  In all empirical studies,  our model based on ResNet-50~\cite{he2016deep} backbone is pretrained for 100 epochs on ImageNet~\cite{deng2009imagenet}. We report all results using the linear evaluation protocol on ImageNet~\cite{chen2021exploring}.  A supervised linear classifier is trained based on the frozen features from the pre-trained model and the number of training epochs   is set to 50. Other training  settings of the linear evaluation are kept the same as~\cite{chen2021exploring}. 
\subsection{ Batch Size}
\label{detail:collapse}
Empirical experiments are conducted to evaluate  the performance of  our method with different batch sizes. We compare MVEB with SimCLR~\cite{DBLP:conf/icml/ChenK0H20}, SimSiam~\cite{chen2021exploring} and VICReg~\cite{DBLP:conf/iclr/BardesPL22}.
We use a symmetric Siamese  network without a predictor network, a momentum encoder, and a stop-gradient operation.  The batch size is set to a range from 128 to 4096. The projector network consists of three linear layers,  each with the output dimensionality set to 8194. We apply BN and ReLU after the first two layers. SGD is used as the optimizer.  The weight decay and the momentum are set to 1e-4 and 0.9, respectively. The basic learning rate is 0.05, scaled with the batch size and divided by 256, and the loss function coefficient $\beta$ is set to 0.01.

The results are reported in Table \ref{table:bs}. MVEB works well over a wide range of batch-size settings. We can observe that the top-1 accuracy of MVEB increases as the  batch size increases. When the batch size varies from 512 to 4096, the accuracies of MVEB are similar. Compared with SimCLR, SimSiam, and VICReg, our MVEB outperforms them by a large margin with different batch sizes. 

Although  MVEB adopts  the Siamese network with direct weight-sharing similar to SimCLR, MVEB can work well  without the requirement of a large batch (e.g., 4096). The behavior of MVEB is also different from SimSiam and VICReg,  both  of which, with the large batch size 4096, drop significantly in accuracy. Moreover, SimSiam relies on the predictor network and  the stop-gradient operation.  Although SimSiam is effective with small bath sizes, it is not well understood, and hard to interpret its architectural tricks~\cite{DBLP:conf/iclr/BardesPL22}. In contrast, MVEB can work well without architectural tricks and the requirement for a large batch size.

\begin{table}[!t]
    \centering
    \caption{Top-1 accuracies (\%) of linear classification on ImageNet of SSL methods pretrained with different batch sizes.}
     \scalebox{1.}{
    \begin{tabular}{@{}lllllll@{}}
    \toprule
    {Batch size}  & 128  & 256  & 512  & 1024 & 2048 & 4096  \\
    \midrule
    SimCLR~\cite{DBLP:conf/icml/ChenK0H20} & - & 57.5 & 60.7 & 62.8 & 64.0 & 64.6\\ 
    SimSiam~\cite{chen2021exploring}                  & 67.3 & 68.1 & 68.1 & 68.0 & 67.9 & 64.0  \\
    VICReg ~\cite{DBLP:conf/iclr/BardesPL22}                     & 67.3 & 67.9 & 68.2 & 68.3 & 68.6 & 67.8 \\
    \midrule
    MVEB (ours)              & \textbf{67.8}    & \textbf{68.5} & \textbf{68.8} & \textbf{68.9} & \textbf{68.9} &\textbf{69.0}  \\
    \bottomrule
    \end{tabular}}
    
    \vspace{-1.5 em}
    
    \label{table:bs}
\end{table}

\subsection{Target Branch Type} 
 The self-supervised learning methods with Siamese  networks adopt different types of the target branch. We select two common types for it: weight-sharing and momentum-update. In SimCLR~\cite{DBLP:conf/icml/ChenK0H20}, two branches share the same weights and are updated simultaneously, which is referred to as a symmetric network. MoCo~\cite{DBLP:conf/cvpr/He0WXG20}  uses the momentum encoder as the target branch, which performs momentum updates according to the other branch.

We use the same  projector network for both weight-sharing
and momentum-update branches. Specifically, the projector network consists of three linear layers,  each with the output dimensionality set to 8194. We apply BN and ReLU after each of the first two layers. The predictor network is not used in Siamese networks.

\vspace{0.5em}
\noindent\textbf{Weight-Sharing Branch.}
The batch size is set to 1024. Other
configurations are kept the same as the pre-training setting in Section \ref{detail:collapse}.

\vspace{0.5em}
\noindent\textbf{Momentum-Update Branch.} We train 100 epochs with the SGD optimizer. The weight decay and the momentum are set to 1e-4 and 0.9, respectively. 
The basic learning rate is 0.1, scaled with the batch size and divided by 256.
We decrease the learning rate to one-thousandth of it with the cosine decay scheduler after a warm-up period of 5 epochs. 
The loss function coefficient $\beta$ is set to 0.01. The batch size is 1024.
Following the setting of BYOL~\cite{DBLP:conf/nips/GrillSATRBDPGAP20}, we update the target branch with the exponential moving by increasing the average parameters from 0.996 to 1 with the cosine scheduler. 

 \begin{figure}
\centering 
\includegraphics[width=0.45\textwidth]{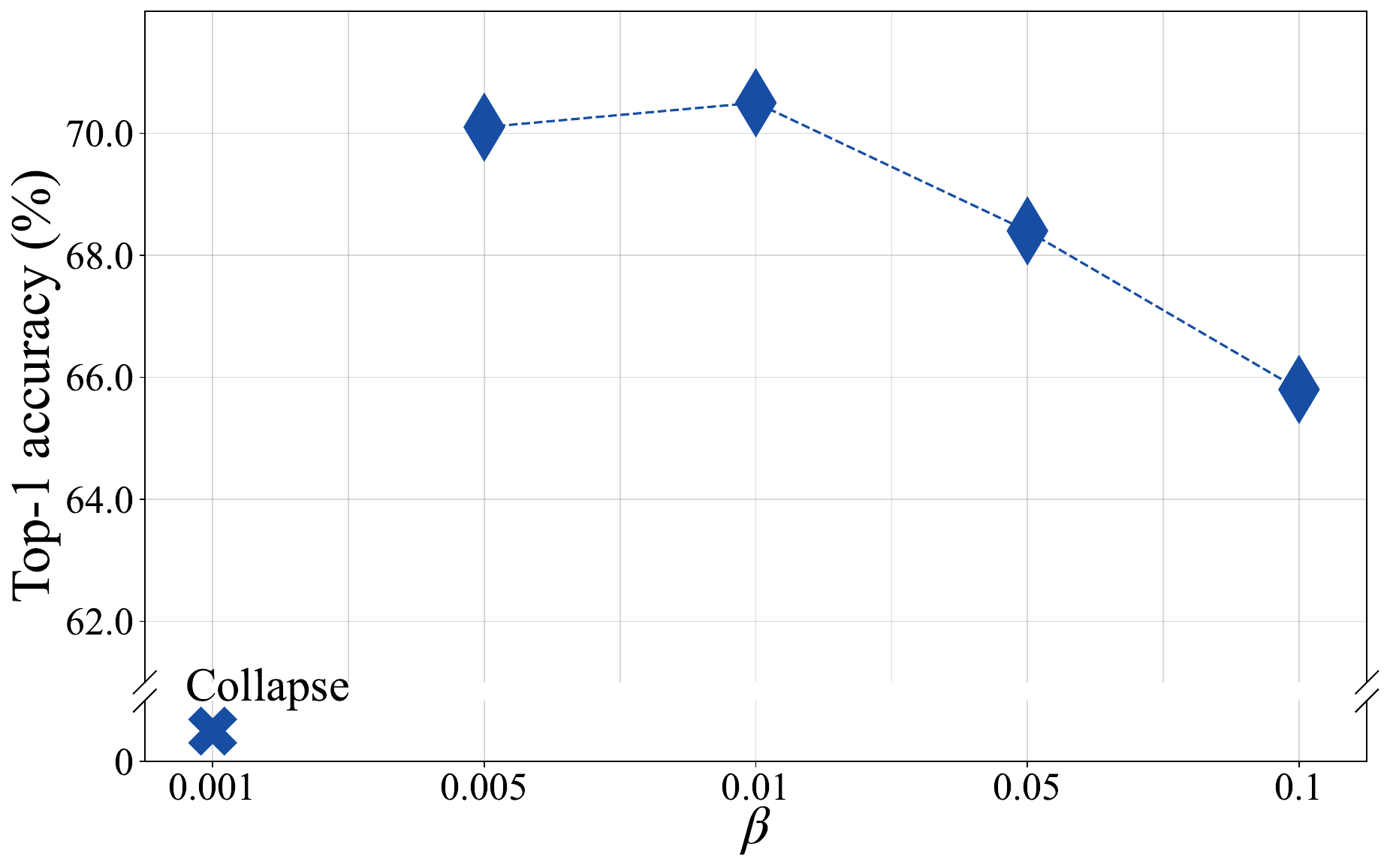}
\caption{Linear classification on ImageNet by MVEB pretrained with different coefficients $\beta$. Collapse means that the accuracy of the linear classification is 0. }
\vspace{-1.3em}
\label{figure:beta}
\end{figure}

We empirically study the effect of these two types  for MVEB.  For the weight-sharing target branch, the linear evaluation on ImageNet is 68.9\%. In contrast, the momentum encoder can achieve  71.2\% with the linear evaluation on ImageNet. This shows that MVEB is more beneficial with the momentum encoder. Hence, we adopt it as the target branch in our  experiments.

\subsection{Loss Balance Coefficient}

The objective function of MVEB in Eq.~(\ref{MVEBf}) consists of two terms, each having different roles. The first term $\mathbb{E}_{ p(\mathbf{z_1}, \mathbf{z_2})} [\mathbf{z_2}^T\mathbf{z_1}]$ learns invariant representation for different views of a sample.  Maximizing the second term $H(\mathbf{z_1})+H(\mathbf{z_1})$  increases the uniformity of the embeddings.

We study its importance and report the performance in Fig.~\ref{figure:beta}.
We can observe that all the representations collapse to a constant vector when $\beta$ is $0.001$. Since $\beta$ approaches zero, the Siamese network suffers from model collapse with the trivial constant representations without maximizing uniformity.
In Eq.~(\ref{MVEB}), $\beta=\frac{1}{(\lambda+1)\kappa }$, where $\kappa$ is a  constant. As shown in Eq.~(\ref{MIB}), $\lambda$ is the coefficient to balance the optimization of $I_\phi\left(\mathbf{z_1} ; \mathbf{v_1} | \mathbf{v_2}\right)$ and $I_\phi\left(\mathbf{z_1} ; \mathbf{v_2}\right)$.
When increasing the value of $\beta$, which decreases $\lambda$, the objective pays more attention to   maximizing $I_\phi\left(\mathbf{z_1}; \mathbf{v_2}\right)$ to keep the information relevant to $\mathbf{v_2}$. However, the performance of the model degrades when  $\beta$ is greater than $0.01$.  
This is because 
increasing $\beta$
to make $\lambda$  smaller than a threshold does not eliminate superfluous information effectively, which hurts the performance of downstream tasks.
\begin{table}
\caption{\wenlj{Top-1 accuracies (\%) of linear classification on ImageNet~\cite{deng2009imagenet} with ViT-s and ResNet-50. The bold entries denote the best. } }	
\begin{center}
\tablestyle{9pt}{1.2}
\begin{adjustbox}{max width=0.48\textwidth}
\begin{tabular}{lcc}
\shline
\diagbox{Method}{Top-1}{Backbone}  & ResNet-50 (23M) &  ViT-s (21M) \\
\hline
SimCLR \cite{DBLP:conf/icml/ChenK0H20}   & 69.3   & 69.0\\
SwAV~\cite{DBLP:conf/nips/CaronMMGBJ20} & 71.8  & 67.1\\
BYOL \cite{DBLP:conf/nips/GrillSATRBDPGAP20} &
 74.3 &71.0 \\
MoCo V3 \cite{DBLP:conf/iccv/ChenXH21}&73.8 &
72.5 \\
DINO \cite{caron2021emerging}& - &70.9\\
MAE\cite{DBLP:conf/cvpr/HeCXLDG22}& - &68.2\\
\hline 
MVEB (ours)& \textbf{74.6} & \textbf{73.4}
\\			
   \shline
	\end{tabular}
	\end{adjustbox}
	\end{center}
	\label{tab:dfb}
 \vspace{-2. em}
\end{table}

\wenlj{\subsection{Generalization Across Different Backbones}
In this section, we evaluate the capability of our approach to generalize effectively across both ViTs and ConvNets. 
We follow the experiment setting on ImageNet in \cite{DBLP:conf/iccv/ChenXH21}, and compare MVEB with previous popular SSL methods based on the Siamese network and MAE\cite{DBLP:conf/cvpr/HeCXLDG22} across both ViTs and
ConvNets in Table  \ref{tab:dfb}.
The results indicate that MVEB demonstrates competitive performance with both  ViTs and ConvNets. }

\subsection{Pretraining Efficiency}
\label{PE}

\wenljx{To evaluate the pretraining efficacy of the MVEB approach, we conduct a comparative analysis against two established methods: BYOL and Barlow Twins. 
To ensure an equitable evaluation, we standardize the experimental setup by employing a Resnet50 architecture as the underlying backbone. We assess the pretraining duration required by each method to complete 1000 epochs with a batch size of 4096 on ImageNet.
The experimental configurations for BYOL and Barlow Twins are aligned with the parameters detailed in their respective seminal papers \cite{DBLP:conf/nips/GrillSATRBDPGAP20} and \cite{zbontar2021barlow}.

MVEB takes approximately 81 hours to distribute on 32 NVIDIA V100 GPUs.  The reimplementations of BYOL and Barlow Twins take approximately 89 and 80 hours on the same hardware with the same setting, respectively.
Hence, the pretraining cost of MVEB is similar to those of BYOL and Barlow Twins.}

\wenlj{\subsection{Comparison of MVEB with MVIB  }

The work in \cite{wang2022rethinking} conducts analytical experiments on CIFAR10 to compare MVIB with  SimCLR. The linear evaluation accuracy of SimCLR is 85.76\%, while the linear evaluation accuracy of MVIB is 86.2\%. This indicates that MVIB cannot improve its performance much compared with SimCLR.  We follow the same experiment setting as \cite{wang2022rethinking} on CIFAR10 to compare MVEB with MVIB.  The linear evaluation accuracy of MVEB is 90.42\%. This accuracy surpasses that of MVIB. Besides, MVEB can be directly applied to the Siamese network, but MVIB  requires an additional stochastic net to obtain the feature distribution as shown in Fig. \ref{figure:MVIB}.}

%% file: doc/conclusion.tex
\section{Discussion and Conclusion}
\label{ss:dc}

The core of self-supervised learning is that the learned representation can generalize well to downstream  tasks. The  minimal sufficient representation can improve the generalization. We propose the multi-view entropy bottleneck (MVEB), a novel pretext task to learn  the  minimal sufficient representation. It can be further simplified to maximizing both the agreement between the embeddings of the two views of a sample and the differential entropy of
the embedding distribution.  We  present the score-based entropy estimator with the von Mises-Fisher kernel to approximate the gradient of the
differential entropy. This estimator can be used to
maximize the differential entropy to prevent collapse efficiently.
Extensive experiments show that MVEB generalizes well across various downstream tasks and establishes new state-of-the-art results.

\textbf{Limitation.} Exiting self-supervised methods with Siamese networks are based on the common assumption in multi-view learning: either view is (approximately) sufficient for the prediction of downstream tasks and contains the same task-relevant information. Hence, the non-shared task-relevant information between the views can be ignored. Our experiment results also verify
that the minimal sufficient representation can improve the
generalization for downstream tasks. If the discrepancy between the two views is too large,  the non-shared task-relevant information cannot  be ignored. In other words,  either view is not sufficient for the prediction of downstream tasks. Thus either view cannot be regarded as a supervised signal to extract task-relevant information and eliminate the  superfluous information in Siamese networks. 
How to overcome this problem is future work.

%% file: doc/sup.tex
\section*{Appendix}
\subsection*{A Stein Gradient Estimator}
\label{sec:stein}
A promising method for estimating the score of an implicit distribution is the  Stein gradient estimator proposed in \cite{DBLP:conf/iclr/LiT18}. Here we briefly describe it.

Assume  $\mathbf{x}$ is supported on $\mathcal{X} \subseteq \mathbb{R}^{d}$. Let $q(\mathbf{x})$ be a continuously differentiable  probability density function, and $\boldsymbol{h}(\mathbf{x})=\left[h_{1}(\mathbf{x}), h_{2}(\mathbf{x}), \ldots, h_{d^{\prime}}(\mathbf{x})\right]^{T}$ be a $d^{\prime}$-dimensional differentiable vector function, where the following boundary condition is satisfied:
\begin{align}
q(\mathbf{x})\boldsymbol{h}(\mathbf{x})=\mathbf{0}, \forall \mathbf{x} \in \partial \mathcal{X}& \text { if } \mathcal{X} \text { is compact}, \text {or}\nonumber \\ 
&\lim _{\mathbf{x} \rightarrow \infty} q(\mathbf{x}) \boldsymbol{h}(\mathbf{x})=\mathbf{0} \text { if } \mathcal{X}=\mathbb{R}^{d}.
\end{align}
We call $\boldsymbol{h}(\mathbf{x})$ the Stein class of $q(\mathbf{x}) $ if the above condition holds.
Then the following Stein's identity can be derived by using integration by parts:
\begin{align}\label{steinid}
\mathbb{E}_{q}\left[\boldsymbol{h}(\mathbf{x}) [\nabla_{\mathbf{x}} \log q(\mathbf{x})]^{T}+\nabla_{\mathbf{x}} \boldsymbol{h}(\mathbf{x})\right]=\mathbf{0},
\end{align}
where $\nabla_{\mathbf{x}} \log q(\mathbf{x})$ is the score of $ q(\mathbf{x})$.

The Monte Carlo method can be adopted to estimate the expectation in Eq.~(\ref{steinid}), which establishes a connection between samples from $q(\mathbf{x})$ and the score $\nabla_{\mathbf{x}} \log q(\mathbf{x})$. Let $\mathbf{x}^{1: M}$ be $M$ i.i.d. samples drawn from $q(\mathbf{x})$. Monte Carlo sampling shows:
\begin{align}\label{MCapprom}
-\frac{1}{M} \mathbf{H G} \approx \overline{\nabla_{\mathbf{x}} \boldsymbol{h}} ,
\end{align}
where $\mathbf{H}=\left[\boldsymbol{h}\left(\mathbf{x}^1\right), \cdots, \boldsymbol{h}\left(\mathbf{x}^M\right)\right] \in \mathbb{R}^{d^{\prime} \times M}$, $\mathbf{G}=\left[\nabla_{\mathbf{x}^1} \log q\left(\mathbf{x}^1\right), \cdots, \nabla_{\mathbf{x}^M} \log q\left(\mathbf{x}^M\right)\right]^{T} \in \mathbb{R}^{M \times d}$, $\overline{\nabla_{\mathbf{x}} \boldsymbol{h}}=\frac{1}{M} \sum_{m=1}^M \nabla_{\mathbf{x}^m} \boldsymbol{h}\left(\mathbf{x}^m\right) \in \mathbb{R}^{d^{\prime} \times d}$ and $\nabla_{\mathbf{x}^m}\boldsymbol{h}\left(\mathbf{x}^m\right)=\left[\nabla_{\mathbf{x}^m} h_1\left(\mathbf{x}^m\right), \ldots, \nabla_{\mathbf{x}^m}h_{d^{\prime}}\left(\mathbf{x}^m\right)\right]^{T} \in \mathbb{R}^{d^{\prime}\times d}$. 
This motivates the ridge regression problem:
\begin{align}\label{ridgeregression}
\underset{\hat{\mathbf{G}} \in \mathbb{R}^{M \times d}}{\operatorname{argmin}}\left\|\overline{\nabla_{\mathbf{x}} \boldsymbol{h}}+\frac{1}{M} \mathbf{H} \hat{\mathbf{G}}\right\|_F^2+\frac{\eta}{M^2}\|\hat{\mathbf{G}}\|_F^2,
\end{align}
where $\|\cdot\|_F^2$ denotes the Frobenius norm of a matrix and $\eta>0$ is the regularization coefficient. An analytic solution of Eq.(\ref{ridgeregression}) is:
\begin{align}\label{ste}
\hat{\mathbf{G}}^{\text{Stein}}=-M(\mathbf{K}+\eta \mathbf{I})^{-1} \mathbf{H}^{T} \overline{\nabla_{\mathbf{x}} \boldsymbol{h}},
\end{align}
where $\mathbf{K}=\mathbf{H}^{T} \mathbf{H}$. We rewrite $\mathbf{K}_{ij}=\boldsymbol{h}\left(\mathbf{x}^i\right)^{T} \boldsymbol{h}\left(\mathbf{x}^j\right)$ by defining a positive definite kernel $k: \mathbb{R}^d \times \mathbb{R}^d \rightarrow \mathbb{R}$ so that $\mathbf{K}_{i j}=k\left(\mathbf{x}^i, \mathbf{x}^j\right)$. 
Similarly, we have $ \left(\mathbf{H}^{T} \overline{\nabla_{\mathbf{x}} \boldsymbol{h}}\right)_{i j}=\frac{1}{M} \sum_{m=1}^M \nabla_{x_j^m} k\left(\mathbf{x}^i, \mathbf{x}^m\right)$. In this way, $\hat{\mathbf{G}}^{\text {Stein }}$ and  the estimation of the score $\nabla_{\mathbf{x}} \log q(\mathbf{x})$  can be obtained.

\subsection*{B Proof of  Eq.~(3): $I_\phi\left(\mathbf{z_1} ; \mathbf{v_1} |\mathbf{v_2}\right)=H(\mathbf{z_1}|\mathbf{v_2})-H(\mathbf{z_1}| \mathbf{v_1},\mathbf{v_2})$}

$I_\phi\left(\mathbf{z_1} ; \mathbf{v_1} |\mathbf{v_2}\right)$ is defined as follows:
\begin{align}
\label{o1}
 I_\phi\left(\mathbf{z_1} ; \mathbf{v_1} |\mathbf{v_2}\right)
 =\int p_\phi\left(\mathbf{z_1} , \mathbf{v_1} |\mathbf{v_2}\right)\log p_\phi\left(\mathbf{z_1} , \mathbf{v_1} |\mathbf{v_2}\right) d\mathbf{z_1}d\mathbf{v_1}&\nonumber\\
 -\int p_\phi\left(\mathbf{z_1} , \mathbf{v_1} |\mathbf{v_2}\right)\log p_\phi\left(\mathbf{z_1}  |\mathbf{v_2}\right) p_\phi\left(\mathbf{v_1} |\mathbf{v_2}\right)d\mathbf{z_1}d\mathbf{v_1}&
\end{align}
The first term on the right-hand side of Eq. \ref{o1} can be decomposed as follows:
\begin{align}
 \int p_\phi\left(\mathbf{z_1} , \mathbf{v_1} |\mathbf{v_2}\right)\log p_\phi\left(\mathbf{z_1} |\mathbf{v_1}, \mathbf{v_2}\right) d\mathbf{z_1}d\mathbf{v_1}\nonumber\\+\int p_\phi\left(\mathbf{z_1} , \mathbf{v_1} |\mathbf{v_2}\right)\log p_\phi\left(\mathbf{v_1}|\mathbf{v_2}\right) d\mathbf{z_1}d\mathbf{v_1},
\end{align}
and the second term  can be decomposed as:
\begin{align}
\int p_\phi\left(\mathbf{z_1} , \mathbf{v_1} |\mathbf{v_2}\right)\log p_\phi\left(\mathbf{z_1}  |\mathbf{v_2}\right) d\mathbf{z_1}d\mathbf{v_1}\nonumber\\+
\int p_\phi\left(\mathbf{z_1} , \mathbf{v_1} |\mathbf{v_2}\right)\log p_\phi\left(\mathbf{v_1} |\mathbf{v_2}\right)d\mathbf{z_1}d\mathbf{v_1}.
\end{align}
Hence, $I_\phi\left(\mathbf{z_1} ; \mathbf{v_1} |\mathbf{v_2}\right)$ is represented as:
\begin{align}
\label{o2}
 I_\phi\left(\mathbf{z_1} ; \mathbf{v_1} |\mathbf{v_2}\right)
 = \int p_\phi\left(\mathbf{z_1} , \mathbf{v_1} |\mathbf{v_2}\right)\log p_\phi\left(\mathbf{z_1} |\mathbf{v_1}, \mathbf{v_2}\right) d\mathbf{z_1}d\mathbf{v_1}&\nonumber\\
 -\int p_\phi\left(\mathbf{z_1} , \mathbf{v_1} |\mathbf{v_2}\right)\log p_\phi\left(\mathbf{z_1}  |\mathbf{v_2}\right) d\mathbf{z_1}d\mathbf{v_1}&\nonumber\\
\end{align}
The first term on the right-hand side of Eq. \ref{o2} can be rewritten as:
\begin{align}
 \int p_\phi\left(\mathbf{z_1} , \mathbf{v_1} |\mathbf{v_2}\right)\log p_\phi\left(\mathbf{z_1} |\mathbf{v_1}, \mathbf{v_2}\right) d\mathbf{z_1}d\mathbf{v_1}=-H(\mathbf{z_1}| \mathbf{v_1},\mathbf{v_2}),
\end{align}
and the  second term can be derived as :
\begin{align}
& -\int p_\phi\left(\mathbf{z_1} , \mathbf{v_1} |\mathbf{v_2}\right)\log p_\phi\left(\mathbf{z_1}  |\mathbf{v_2}\right) d\mathbf{z_1}d\mathbf{v_1}=\nonumber\\
 &-\int p_\phi\left(\mathbf{z_1} |\mathbf{v_2}\right)\log p_\phi\left(\mathbf{z_1}  |\mathbf{v_2}\right) d\mathbf{z_1}=H(\mathbf{z_1}|\mathbf{v_2}).\nonumber
\end{align}

Finally, we obtain:
\begin{align}
 I_\phi\left(\mathbf{z_1} ; \mathbf{v_1} |\mathbf{v_2}\right)=H(\mathbf{z_1}|\mathbf{v_2})-H(\mathbf{z_1}| \mathbf{v_1},\mathbf{v_2}).   
\end{align}

\subsection*{C Proof of  Eq. (4): $I_\phi\left(\mathbf{z_1} ; \mathbf{v_1} \right)=H(\mathbf{z_1})-H(\mathbf{z_1}| \mathbf{v_1})$}

$I_\phi\left(\mathbf{z_1} ; \mathbf{v_1} \right)$ is defined as follows:
\begin{align}
\label{o21}
 I_\phi\left(\mathbf{z_1} ; \mathbf{v_1} \right)
 =\int p_\phi\left(\mathbf{z_1} , \mathbf{v_1} \right)\log p_\phi\left(\mathbf{z_1} , \mathbf{v_1} \right) d\mathbf{z_1}d\mathbf{v_1}&\nonumber\\
 -\int p_\phi\left(\mathbf{z_1} , \mathbf{v_1} \right)\log p_\phi\left(\mathbf{z_1}  \right) p_\phi\left(\mathbf{v_1} \right)d\mathbf{z_1}d\mathbf{v_1}.&
\end{align}
The first term on the right-hand side of Eq. \ref{o21} can be decomposed as follows:
\begin{align}
 \int p_\phi\left(\mathbf{z_1} , \mathbf{v_1} \right)\log p_\phi\left(\mathbf{z_1} |\mathbf{v_1}\right) d\mathbf{z_1}d\mathbf{v_1}\nonumber\\+\int p_\phi\left(\mathbf{z_1} , \mathbf{v_1} \right)\log p_\phi\left(\mathbf{v_1}\right) d\mathbf{z_1}d\mathbf{v_1},
\end{align}
and the second term  can be decomposed as:
\begin{align}
\int p_\phi\left(\mathbf{z_1} , \mathbf{v_1} \right)\log p_\phi\left(\mathbf{z_1}  \right) d\mathbf{z_1}d\mathbf{v_1}\nonumber\\+
\int p_\phi\left(\mathbf{z_1} , \mathbf{v_1} \right)\log p_\phi\left(\mathbf{v_1} \right)d\mathbf{z_1}d\mathbf{v_1}.
\end{align}
Hence, $I_\phi\left(\mathbf{z_1} ; \mathbf{v_1} \right)$ is represented as:
\begin{align}
\label{o22}
 I_\phi\left(\mathbf{z_1} ; \mathbf{v_1} \right)
 = \int p_\phi\left(\mathbf{z_1} , \mathbf{v_1} \right)\log p_\phi\left(\mathbf{z_1} |\mathbf{v_1}\right) d\mathbf{z_1}d\mathbf{v_1}&\nonumber\\
 -\int p_\phi\left(\mathbf{z_1} , \mathbf{v_1} \right)\log p_\phi\left(\mathbf{z_1} \right) d\mathbf{z_1}d\mathbf{v_1}.&\nonumber\\
\end{align}
The first term on the right-hand side of Eq. \ref{o22} can be rewritten as:
\begin{align}
 \int p_\phi\left(\mathbf{z_1} , \mathbf{v_1} \right)\log p_\phi\left(\mathbf{z_1} |\mathbf{v_1}\right) d\mathbf{z_1}d\mathbf{v_1}=-H(\mathbf{z_1}| \mathbf{v_1}),
\end{align}
and the second term can be derived as :
\begin{align}
& -\int p_\phi\left(\mathbf{z_1} , \mathbf{v_1} \right)\log p_\phi\left(\mathbf{z_1}  \right) d\mathbf{z_1}d\mathbf{v_1}=\nonumber\\
 &-\int p_\phi\left(\mathbf{z_1} \right)\log p_\phi\left(\mathbf{z_1} \right) d\mathbf{z_1}=H(\mathbf{z_1}).\nonumber
\end{align}

Finally, we obtain:
\begin{align}
 I_\phi\left(\mathbf{z_1} ; \mathbf{v_1} \right)=H(\mathbf{z_1}-H(\mathbf{z_1}| \mathbf{v_1}).   
\end{align}